\documentclass{article}

     \PassOptionsToPackage{numbers, compress}{natbib}

     \usepackage[preprint]{neurips_2020}

\usepackage[utf8]{inputenc} 
\usepackage[T1]{fontenc}    
\usepackage{hyperref}       
\usepackage{url}            
\usepackage{booktabs}       
\usepackage{amsfonts}       
\usepackage{nicefrac}       
\usepackage{microtype}      

\usepackage{algorithm}
\usepackage{algcompatible}
\usepackage{amsfonts}
\usepackage{amsmath}
\usepackage{graphicx}
\usepackage{subcaption}
\usepackage{wrapfig}
\usepackage{arydshln}

\usepackage{xcolor}

\newcommand{\vect}[1]{{\boldsymbol{#1}}}

\newcommand{\Real}{\mathbb{R} }
\newcommand{\E}{\mathbb{E} }

\newcommand{\target}{\textrm{target}}
\newcommand{\prop}{\textrm{prop}}
\newcommand{\intd}{\textrm{d} }
\newcommand{\KL}{D_{\textrm{KL}}}

\title{Learning the Solution Manifold in Optimization and Its Application in Motion Planning}

\author{%
  Takayuki~Osa\textsuperscript{1,2}
    \\
  \textsuperscript{1}Kyushu Institute of Technology\\
  \textsuperscript{2}RIKEN Center for Advanced Intelligence Project\\
  \texttt{osa@brain.kyutech.ac.jp} \\
}

\begin{document}

\maketitle

\begin{abstract}
	Optimization is an essential component for solving problems in wide-ranging fields.
	Ideally, the objective function should be designed such that the solution is unique and the optimization problem can be  solved stably.
	However, the objective function used in a practical application is usually non-convex, and sometimes it even has an infinite set of solutions.
	To address this issue, we propose to learn the solution manifold in optimization.
	We train a model conditioned on the latent variable such that the model represents an infinite set of solutions.
	In our framework, we reduce this problem to density estimation by using importance sampling, and the latent representation of the solutions is learned by maximizing the variational lower bound.
	We apply the proposed algorithm to motion-planning problems, which involve the optimization of high-dimensional parameters. The experimental results indicate that the solution manifold can be learned with the proposed algorithm, and the trained model represents an infinite set of homotopic solutions for motion-planning problems.
\end{abstract}

\section{Introduction}
Optimization is an essential component for problem-solving in various fields, such as machine learning~\citep{Bishop06} and robotics~\citep{Siciliano09}.
The objective function, ideally, should be designed in such a way that the solution is unique and the optimization problem can be solved stably.
If we can formulate our problem as convex optimization, we can leverage techniques established in previous studies~\citep{Boyd04}.
However, the objective functions used in practical applications are usually non-convex.
For instance, various optimization methods has been leveraged in motion planning in robotics~\citep{,Khatib86,Zucker13,Schulman14} wherein the objective function is known to be non-convex. 
Moreover, there often exist a set of solutions in motion-planning problems as indicated by~\cite{Jaillet08,Orthey20}. 
It is possible to specify the unique solution by posing additional constraints, but it is not practical for users who are not experts on optimization and/or robotics.
For this reason, a generic objective function is often used in motion planning, which is applicable for various tasks but often has many solutions.
However, existing methods are limited to generating a fixed number of solutions, and pose complications when used to generate a new “middle” solution out of the two obtained solutions.

To address this issue, we propose to learn the solution manifold in optimization problems wherein the objective function has an infinite set of optimal points.
Our main idea is to train a model conditioned on the latent variable such that the model represents an infinite set of solutions.
We reduce the problem of learning the solution manifold to that of density estimation using the importance sampling technique.
In this study, we refer to our algorithm as \textit{Learning Solution Manifold in Optimization}~(LSMO).
The objective function with an infinite set of optimal points and how the LSMO captures them are shown in Figure~1. 
The model conditioned on the latent variable is trained by maximizing the variational lower bound as in a variational autoencoder~(VAE)~\citep{Kingma14}.

Our contribution is an algorithm to train the model that represents the distribution of optimal points in optimization.
We evaluate LSMO for maximizing the test functions that have an infinite set of optimal points. We also apply LSMO to motion-planning tasks for a robotic manipulator, which involves hundreds of parameters to be optimized.
Two trajectories are called \textit{homotopic} when one can be continuously deformed into the other ~\citep{Jaillet08,Hatcher02}. 
We empirically show that LSMO can learn an infinite set of homotopic trajectories in motion-planning problems.


\section{Optimization via Inference with Importance Sampling}
\label{sec:prob}
\begin{figure}
	\centering
	\begin{subfigure}[t]{0.3\columnwidth}
		\centering
		\includegraphics[width=\textwidth]{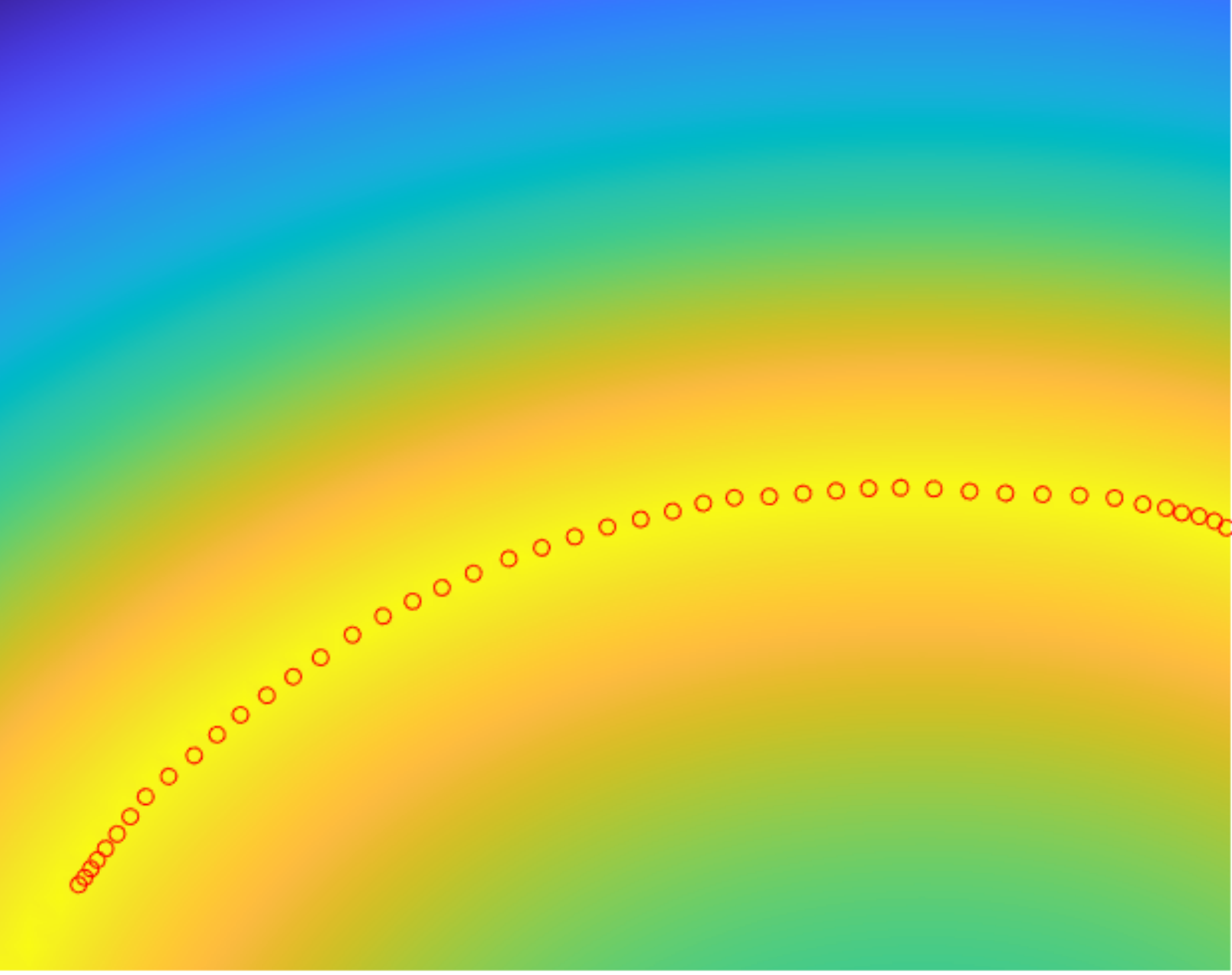}
		\caption{Visualization of the objective function and solutions obtained with LSMO. }
	\end{subfigure}
	\begin{subfigure}[t]{0.3\columnwidth}
		\centering
		\includegraphics[width=\textwidth]{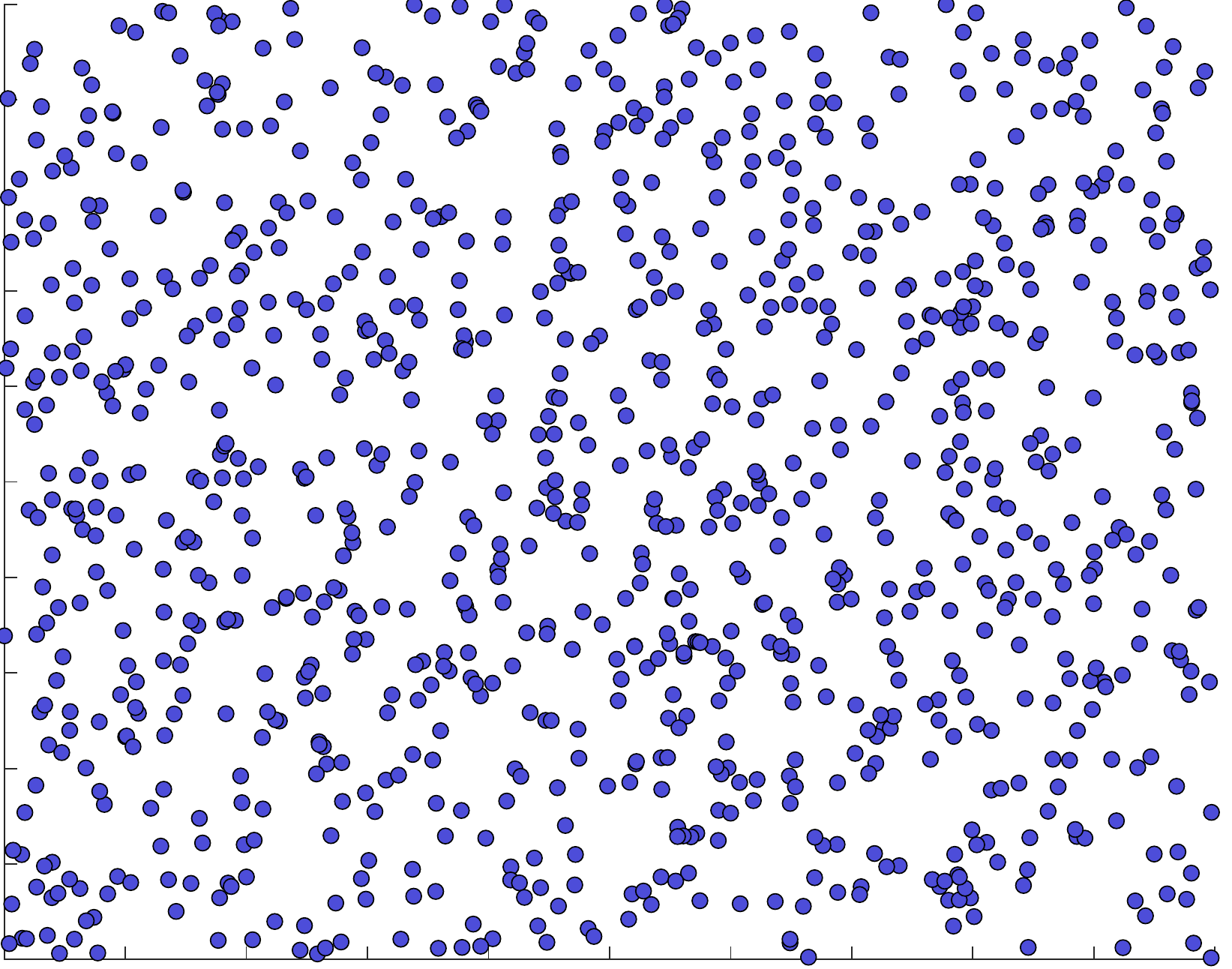}
		\caption{Samples drawn from the uniform distribution.}
	\end{subfigure}
	\begin{subfigure}[t]{0.3\columnwidth}
		\centering
		\includegraphics[width=\textwidth]{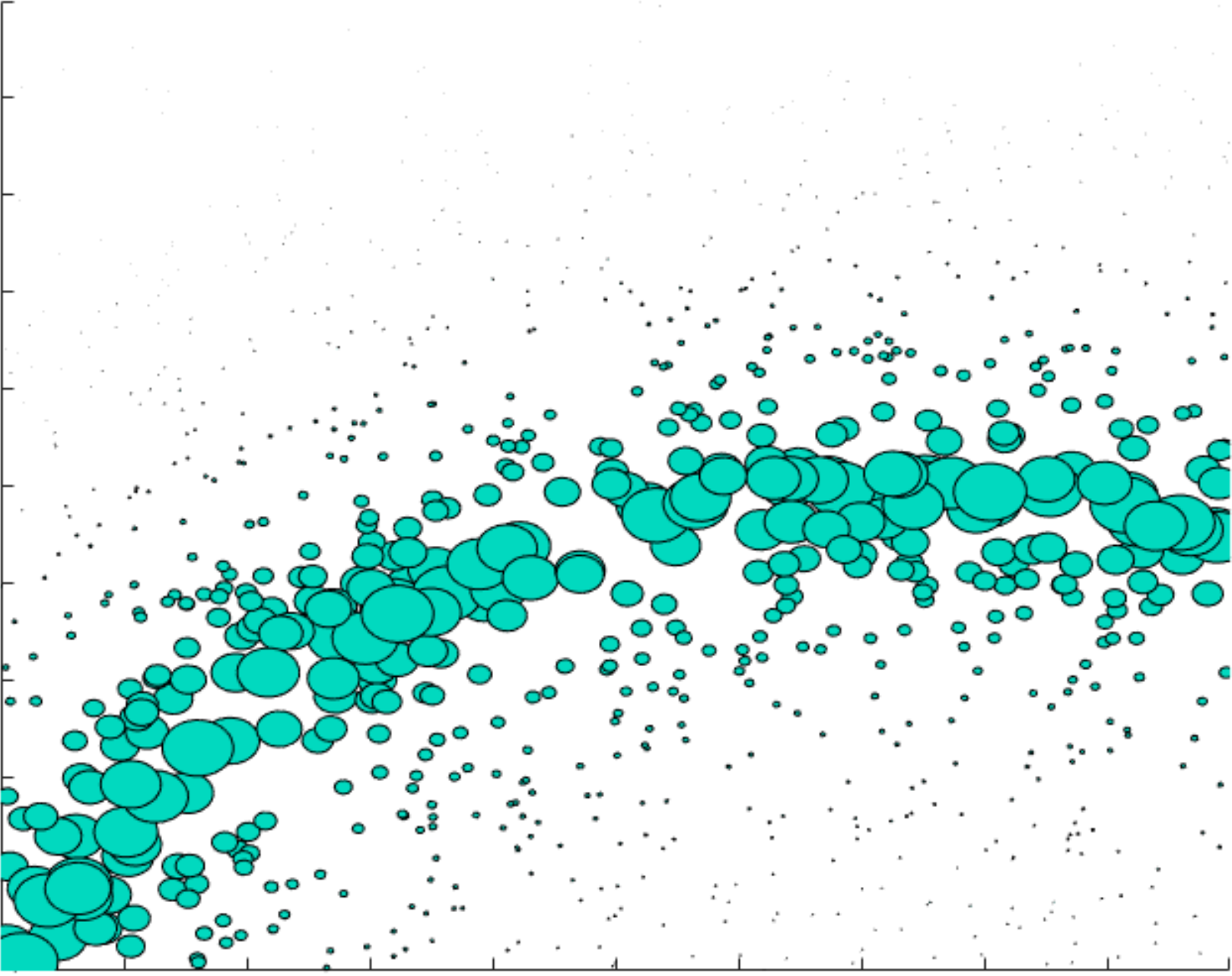}
		\caption{The same samples in (b) with the scaling based on the importance weight.}
	\end{subfigure}
	\caption{Example of learning the solution manifold.  (a) shows an objective function that has an infinite set of solutions. The warmer color represents the higher return in (a).  Our method trains $p_{\vect{\theta}}(\vect{x}|\vect{z})$ that outputs the samples indicated by red circles using different values of $\vect{z}$. (b) shows samples drawn from a uniform distribution. In (c), the same samples as those in (b) are shown, but samples with higher importance are drawn as larger circles. We train $p_{\vect{\theta}}(\vect{x}|\vect{z})$ with this importance weight.
	} 
	\label{fig:approach}
\end{figure}

\paragraph{Problem formulation}
We consider the problem of maximizing the objective function, which is formulated as
\begin{align}
\vect{x}^* = \arg \max_{\vect{x}} R(\vect{x}),
\end{align}
where $\vect{x} \in \mathcal{X}$ is a data point and $R(\vect{x}): \mathcal{X} \mapsto \Real$ is the objective function.
In this study, we are particularly interested in the problems where there exist multiple, and possibly an infinite number of solutions.
For example, the objective function shown in Figure~\ref{fig:approach}(a) has an infinite number of solutions.
The goal of our study is to learn the solution manifold when optimizing such objective functions.
Instead of finding a single solution, we aim to train a model that represents the distribution of optimal points $p_{\vect{\theta}}(\vect{x})$ parameterized with a vector $\vect{\theta}$ given by  
\begin{align}
p_{\vect{\theta}}(\vect{x})= \int p_{\vect{\theta}}(\vect{x} | \vect{z} )p(\vect{z}) \textrm{d}\vect{z},
\end{align}
where $\vect{z}$ is the latent variable.
We train the model $p_{\vect{\theta}}(\vect{x} | \vect{z} )$ by maximizing the surrogate objective function
\begin{align}
J(\vect{\theta}) = \E_{\vect{x} \sim p_{\vect{\theta}}(\vect{x})} [ f\big(R(\vect{x})\big) ],
\label{eq:surrogate_obj}
\end{align}
where $f(\cdot)$ is monotonically increasing and $f(x) \geq 0$ for any $x \in \Real$.
Since $f(\cdot)$ is monotonically increasing, maximizing $R(\vect{x})$ is equivalent to maximizing $f(R(\vect{x}))$. Therefore, $f(\cdot)$ can be interpreted as a shaping function.

\paragraph{Inference with importance sampling}
To address the problem formulated in the previous paragraph, we consider a distribution of the form
\begin{equation}
p^{\target}( \vect{x} ) = \frac{ f \left( R(\vect{x}) \right)}{ Z },
\label{eq:target}
\end{equation}
where $Z$ is a partition function.
When $p^{\target}( \vect{x} )$ is followed, a sample with a higher score is drawn with a higher probability. 
Therefore, the modes of the density induced by $p^{\target}( \vect{x} )$ correspond to the modes of $R(\vect{x})$.
Thus, we reduce the problem of finding points that maximize the objective function $R(\vect{x})$ to that of estimating the density induced by  $p^{\target}( \vect{x} )$, which is formulated as 
\begin{align}
\min_{\vect{\theta}} D_{\textrm{KL}}( p^{\target}(\vect{x}) || p_{\vect{\theta}}(\vect{x} ) ),
\label{eq:prob_KL}
\end{align}
where $ D_{\textrm{KL}}( p^{*}(\vect{x}) || p_{\vect{\theta}}(\vect{x} ) )$ is the KL divergence. To solve this problem, we employ the importance sampling technique. 
We introduce the importance weight of the data point $\vect{x}_i$ defined by
\begin{align}
W(\vect{x}_i) = \frac{p^{\target}( \vect{x}_i )}{ p_{\prop}(\vect{x}_i) },
\label{eq:weight}
\end{align}
where $p_{\prop}(\vect{x})$ is the proposal distribution.
The KL divergence$ D_{\textrm{KL}}( p^{*}(\vect{x}) || p_{\vect{\theta}}(\vect{x} ) )$ can be rewritten as
\begin{align}
& D_{\textrm{KL}}( p^{\target}(\vect{x}) ||p_{\vect{\theta}}(\vect{x} ) ) 
= \int p^{\target}(\vect{x})\log \frac{p^{\target}(\vect{x})}{ p_{\vect{\theta}}(\vect{x})} \textrm{d}\vect{x} 
= \int W(\vect{x}) p_{\textrm{prop}}(\vect{x})\log \frac{ W(\vect{x}) p_{\textrm{prop}}(\vect{x})}{ p_{\vect{\theta}}(\vect{x})} \textrm{d}\vect{x}.
\label{eq:kl_p}
\end{align}
Therefore, given a set of samples $X=\{ \vect{x}_i \}^N_{i=1}$ drawn from the proposal distribution, the minimizer of $ D_{\textrm{KL}}( p^{\target}(\vect{x}) || p_{\vect{\theta}}(\vect{s}, \vect{\tau} ) )$ is given by the maximizer of the weighted log likelihood:
\begin{align}
L( \vect{\theta} )  = \int  W(\vect{x}) p_{\textrm{prop}}(\vect{x}) \log p_{\vect{\theta}}(\vect{x} ) \textrm{d}\vect{x}
\approx \frac{1}{N} \sum_{ i=1 }^N W(\vect{x}_{i}) \log p_{\vect{\theta}}(\vect{x}_{i} ).
\label{eq:ml}
\end{align}

\paragraph{Connection to the original problem} 

Because $f(x)$ is monotonically increasing and positive for any $x$,  in a manner similar to the results shown by \cite{Dayan97,Kober11}, we can obtain 
\begin{align}
\log J( \vect{\theta}  ) 
& = \log \int p_{\vect{\theta}}( \vect{x}) f\big(R(\vect{x})\big) \textrm{d}\vect{x}
= \log \int p_{\textrm{prop}}( \vect{x}) \frac{ f\big(R(\vect{x})\big)}{ p_{\textrm{prop}}( \vect{x}) } 
p_{\vect{\theta}}( \vect{x}) \textrm{d}\vect{x} \nonumber \\
&\geq  \int p_{\textrm{prop}}( \vect{x}) \frac{ f\big(R(\vect{x})\big)}{ p_{\textrm{prop}}( \vect{x})  } 
\log p_{\pi}( \vect{x}) \textrm{d}\vect{x} 
=   Z \int p_{\textrm{prop}}( \vect{x}) \frac{ f\big(R(\vect{x})\big)}{ p_{\textrm{prop}}( \vect{x} ) Z} 
\log p_{\vect{\theta}}( \vect{x}) \textrm{d}\vect{x} \nonumber \\
&\propto L(  \vect{\theta}).
\label{eq:bound}
\end{align}
Therefore, if we interpret $f(\cdot)$ as a shaping function, minimizing the KL divergence in \eqref{eq:prob_KL} can be interpreted as maximization of the lower bound of the expected value of the shaped objective function.

\section{Learning the Solution Manifold with Importance Sampling}
To train the model $p_{\vect{\theta}}(\vect{x}|\vect{z})$, we leverage the variational lower bound as in VAE~\citep{Kingma14}.
The variational lower bound on the marginal likelihood of data point $i$ is given by
\begin{align}
\log p_{\vect{\theta}}(\vect{x}_i) & \geq \mathcal{L}(\vect{\psi}, \vect{\theta};\vect{x}_i) 
= - \KL\left( q_{\vect{\psi}}(\vect{z}|\vect{x}_i)|| p(\vect{z}) \right) + \E_{\vect{z} \sim q(\vect{z}|\vect{x}_i)} \left[ \log p_{\vect{\theta}}(\vect{x}_i | \vect{z}) \right]
\label{eq:elbo}
\end{align}
This variational lower bound can be used to obtain the lower bound of the objective function in~\eqref{eq:ml}.
Given a set of data points $X=\{\vect{x}_i\}^N_{i=1}$, which are samples collected by a proposal distribution, we train the model by maximizing the following function:
\begin{align}
\mathcal{L}(\vect{\psi}, \vect{\theta}; X) = \sum^{N}_{i=1} W(\vect{x}_i) \left( - \KL\left( q_{\vect{\psi}}(\vect{z}|\vect{x}_i)|| p(\vect{z}) \right) + \log p_{\vect{\theta}}(\vect{x}_i | \vect{z}_i) \right)
\label{eq:vae_obj}
\end{align}
where $\vect{z}_i$ is a sample drawn from the posterior distribution $q_{\vect{\psi}}(\vect{z}|\vect{x}_i)$.
From \eqref{eq:bound} and \eqref{eq:elbo}, we can see that the loss function in \eqref{eq:vae_obj} is the lower bound of the surrogate objective function in \eqref{eq:surrogate_obj}.
The form of our loss function in \eqref{eq:vae_obj} is a simple combination of the return-weighted importance in \eqref{eq:weight}.
In practice, the partition function $Z$ in \eqref{eq:target} is an unknown constant and does not depend on $\vect{\theta}$. 
Therefore, we use $\tilde{W}(\vect{x}) =  \frac{ f \left( R(\vect{x}) \right)}{ p_{\prop}(x) }$ instead of $W(\vect{x})$ in our implementation.

The above discussion indicates that we can leverage various techniques for learning latent representations based on VAE, such as $\beta$-VAE~\citep{Higgins16} and joint VAE~\citep{Dupont18}.
In our implementation, we used the loss function based on joint VAE proposed by \citet{Dupont18}, which is given by
\begin{align}
\tilde{\mathcal{L}}(\vect{\psi}, \vect{\theta}; X) = \sum^{N}_{i=1} \tilde{W}(\vect{x}_i) \ell(\vect{\theta}, \vect{\psi}),
\label{eq:loss}
\end{align}
where $\ell(\vect{\theta}, \vect{\psi})$ is given by
\begin{align}
\ell(\vect{\theta}, \vect{\psi})  =  \log p_{\vect{\theta}} (\vect{x}|\vect{z} )  
- \gamma \left| \KL\big( q_{\vect{\psi}}(\vect{z} | \vect{x}) || p(\vect{z}) \big) - C_{\vect{z}} \right|,
\end{align}
$C_{\vect{z}}$ represents the information capacity, and $\gamma$ is a coefficient.
$\KL\big( q_{\vect{\psi}}(\vect{z} | \vect{x}) || p(\vect{z}) \big)$ is the upper bound of the mutual information between $\vect{z}$ and $\vect{x}$, and \citet{Dupont18} empirically show that the scheduling $C_{\vect{z}}$ encourages the learning disentangled representations.

The proposed algorithm for learning the manifold of solutions is summarized in Algorithm~\ref{alg:LSMO}.
After the first iteration, we can replace the proposal distribution with a distribution induced by injecting noise to the trained model $p_{\vect{\theta}}(\vect{x}) = \int p_{\vect{\theta}}(\vect{x}|\vect{z})p(\vect{z}) \intd \vect{z}$. The details of updating the proposal distribution are provided in the Appendix~A.3.

\begin{algorithm}[t]
	\caption{Learning the solution manifold in optimization (LSMO) }
	\begin{algorithmic}[1]
		\STATEx{\textbf{Input:} objective function $R(\vect{x})$, proposal distribution $p_{\prop}$ 
		}
		\STATE{Generate $N$ synthetic samples $\{\vect{x}_i\}^N_{i=1}$  }
		\STATE{Evaluate the objective function $R(\vect{x}_i)$ for $i=1,\ldots, N$.  }
		\STATE{Compute the weight $W(\vect{x}_i)$ for $i=1,\ldots, N$.  }
		\STATE{Train the neural network by minimizing  $\tilde{\mathcal{L}}(\vect{\psi}, \vect{\theta}; X)$ in \eqref{eq:loss} }
		\STATE{(Optional) Update the proposal distribution and repeat 1-3}
		\STATEx{\textbf{Return:}  $p_{\vect{\theta}}(\vect{x}|\vect{z})$ }
		
	\end{algorithmic}
	\label{alg:LSMO}
\end{algorithm}

\section{Application to Motion Planning for Robotic Manipulator}
We apply the LSMO algorithm to motion planning for a robotic manipulator.
We introduce the formulation of motion planning problems and present how to adapt LSMO for motion planning. 

\paragraph{Problem Formulation} We denote by $\vect{q}_t \in \Real^{D}$ the configuration of a robot manipulator with $D$ degrees of freedom~(DoFs) at time $t$.
Given the start configuration $\vect{q}_0$ and the goal configuration $\vect{q}_T$, the task is to plan a smooth and collision-free trajectory $\vect{\xi} = [\vect{q}_0, \ldots,\vect{q}_T  ] \in \Real^{D \times T} $, which is given by a sequence of robot configurations.
This problem can be formulated as an optimization problem 
\begin{align}
\vect{\xi}^* = \arg \min_{\vect{\xi}} \mathcal{C}(\vect{\xi})
\end{align}
where $\mathcal{C}(\vect{\xi})$ is the cost function that quantifies the quality of a trajectory $\vect{\xi}$.
In LSMO, we train the model $p_{\vect{\theta}}(\vect{\xi}|\vect{z})$ to obtain a set of solutions.

\begin{wrapfigure}{r}{0.3\textwidth}
	\vspace{-0.7cm}
	\begin{center}
		\includegraphics[width=0.28\textwidth]{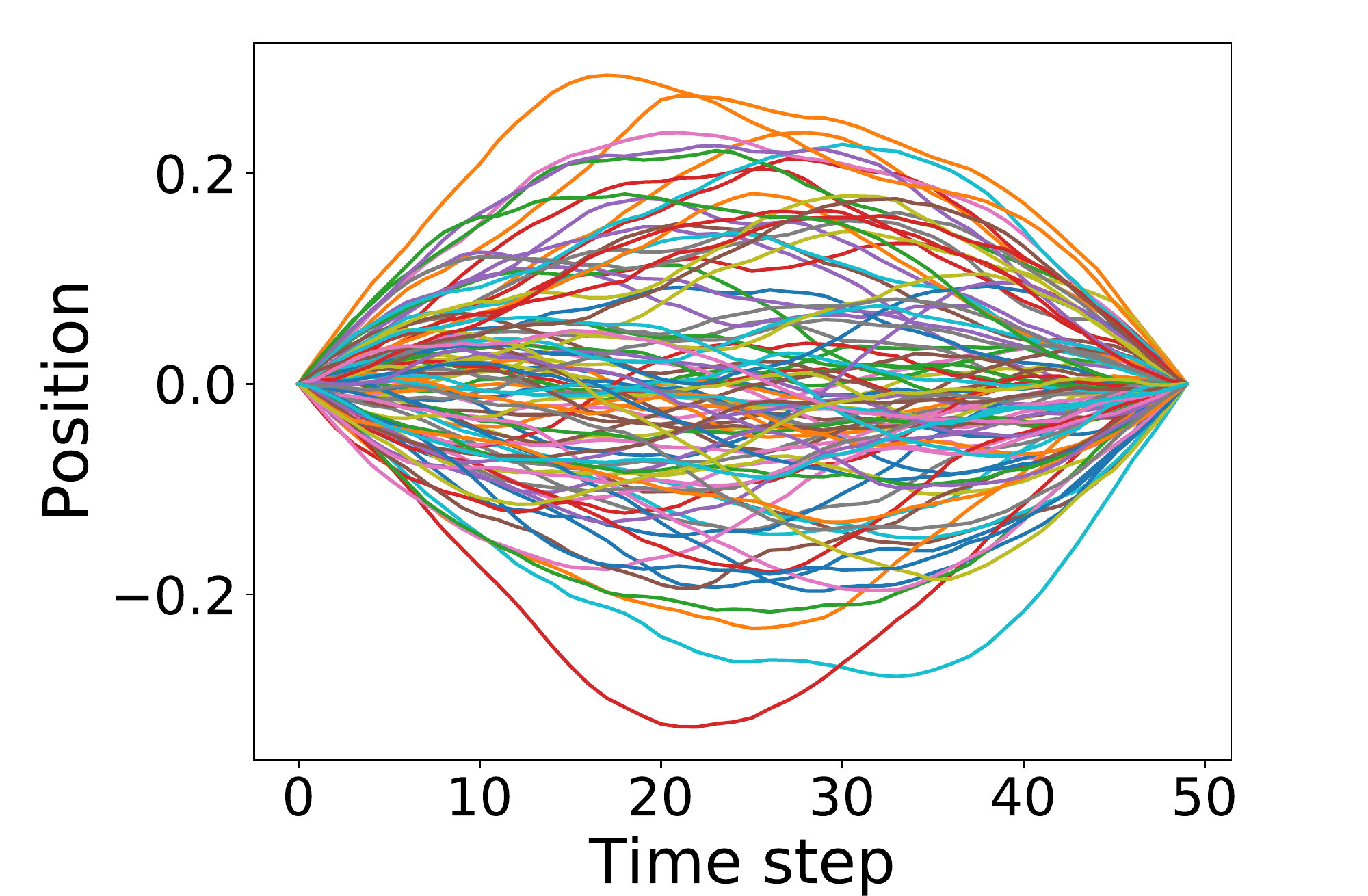}
	\end{center}
	\caption{Visualization of samples drawn from $\beta_{\textrm{traj}}(\vect{\xi})$}
	\vspace{-0.cm}
\end{wrapfigure}

\paragraph{Proposal distribution}
To deal with the exploration of the high-dimensional space of trajectories, it is necessary to use a structured sampling strategy, which is suitable for trajectory optimization. We use the following distribution as a proposal distribution, which is adapted from \cite{Kalakrishnan11}:
\begin{align}
\beta_{\textrm{traj}}(\vect{\xi}) = \mathcal{N}(\vect{\xi}^0, a R)
\label{eq:exploration_traj}
\end{align}	
where $a$ is a constant, $\vect{\xi}^0$ is an initial trajectory, and the covariance matrix $R$ is given by the Moore-Penrose pseudo inverse of the matrix $A$ defined as $A = K^\top K$ and $K$ is a finite difference matrix.
We visualize the trajectories drawn from $\beta_{\textrm{traj}}$ in \eqref{eq:exploration_traj} in the case where the initial trajectory $\vect{\xi}$ is a zero vector. As shown, the trajectories generated from $\beta_{\textrm{traj}}$ start from zero and end with zero. This property of $\beta_{\textrm{traj}}(\vect{\xi})$ is suitable for motion planning problems where the start and goal configurations are given.

\paragraph{Projection onto the Constraint Solution Space}
Although LSMO can learn the manifold of the objective function, there is no guarantee that the output of $p_{\vect{\theta}}(\vect{\xi}|\vect{z})$ satisfies the desired constraint.
For this reason, we fine-tune the output of $p_{\vect{\theta}}(\vect{\xi}|\vect{z})$ by using the covariant Hamiltonian optimization for motion planning~(CHOMP)~\cite{Zucker13}.
The update rule of CHOMP is given by:
\begin{align}
\vect{\xi}^{*} = \arg \min_{\vect{\xi}} \left\{  \mathcal{C}(\vect{\xi}^c) + \vect{g}^{\top} (\vect{\xi} - \vect{\xi}^c) + \frac{\eta}{2} \left\| \vect{\xi} - \vect{\xi}^c \right\|^2_{M}  \right\},
\label{eq:chomp}
\end{align}
where $\vect{g} = \nabla \mathcal{C}(\vect{\xi})$, $\vect{\xi}^c$ is the current plan of the trajectory, 
$\eta$ is a regularization constant, 
and  $\left\| \vect{\xi} \right\|^2_{M}$ is the norm defined by a matrix $M$ as $\left\| \vect{\xi} \right\|^2_{M} = \vect{\xi}^{\top}M\vect{\xi}$.
The third term on RHS in \eqref{eq:chomp} can be interpreted as the trust region, and this term encourages the preservation of the velocity profile of the motion when $M=A$. Using this update rule, we can project the output of $p_{\vect{\theta}}(\vect{x}|\vect{z})$ onto the constraint solution space, while maintaining the velocity profile induced by the latent variable $\vect{z}$.

\begin{algorithm}[t]
	\caption{Motion Planning by Learning the Solution Manifold in Optimization }
	\begin{algorithmic}[1]
		\STATEx{\textbf{Input:} start configuration $\vect{q}_0$, goal configuration $\vect{q}_T$  
		}
		\STATE{Initialize the trajectory, e.g., linear interpolation between $\vect{q}_0$ and $\vect{q}_T$}
		\STATE{Generate $N$ synthetic samples $\{\vect{\xi}_i\}^N_{i=1}$ from $\beta_{\textrm{traj}}(\vect{\xi})$ in \eqref{eq:exploration_traj}  }
		\STATE{Evaluate the objective function $\mathcal{C}(\vect{\xi}_i)$ and compute the weight $W(\vect{\xi}_i)$ for $i=1,\ldots, N$  }
		\STATE{Train the neural network by minimizing  $\mathcal{L}(\vect{\theta}, \vect{\psi})$ in \eqref{eq:loss} }
		\STATE{(Optional) Update the proposal distribution and repeat 1-5.}
		\STATE{Generate  $\vect{\xi}$ with $p_{\vect{\theta}} (\vect{\xi}|\vect{z})$ by specifying the value of $\vect{z}$}
		\STATE{(Optional) Project $\vect{\xi}$ onto the constraint solution space, e.g., using CHOMP \eqref{eq:chomp}  }
		\STATEx{\textbf{Return:}  planned trajectory $\vect{\xi}$ }
		
	\end{algorithmic}
	\label{alg:MPLSM}
\end{algorithm}

\section{Related Work}
\paragraph{Multimodal optimization with black-box optimization methods}
Prior work such as ~\citep{Goldberg87,Deb10,Stoean10,Agrawal14} addressed multimodal optimization using black-box optimization methods, such as CMA-ES~\citep{Hansen96} and the cross-entropy method~(CEM)~\citep{Boer05}.
For example, the study in \cite{Agrawal14} showed that control policies to achieve diverse behaviors can be learned by maximizing the objective function that encodes the diversity of solutions.
Although these black-box optimization methods are applicable to a wide range of problems, the limitation of these methods is that they can learn only a finite set of solutions.

\paragraph{Motion planning methods in robotics} 
A popular class of motion planning methods is optimization-based methods, including CHOMP~\citep{Zucker13}, 
TrajOpt~\citep{Schulman14}, and GPMP~\citep{Mukadam18}.
These methods are designed to determine a single solution, although the objective function is usually non-convex.
A recent study by \cite{osa20} proposed a method, stochastic multimodal trajectory optimization~(SMTO), which finds multiple solutions by learning discrete latent variables based on the objective function.
However, SMTO is limited to learning a finite set of solutions.
Sampling-based methods are also popular in motion planning for robotic systems.
PRM~\citep{Kavraki96,Kavraki98}, RRT~\citep{LaValle01}, and RRT*~\citep{Karaman11} are categorized as such sampling-based methods. 
Studies in \cite{Schmitzberger02,Jaillet08} discussed homotopy and the deformability of trajectories and proposed methods for compactly capturing the variety of trajectories inside a PRM.
Recently, \citet{Orthey20} proposed a framework for finding multiple solutions using a sampling-based method.
However, it is also limited to generate a finite number of solutions. 

\paragraph{Reinforcement learning and imitation learning}
Studies on imitation learning have proposed methods for modeling diverse behaviors by learning latent representations~\citep{Li17,Merel19,Sharma19}.
Likewise, in the field of reinforcement learning~(RL), previous studies have proposed methods for learning the diverse behaviors~\citep{Eysenbach19}.
Hierarchical RL methods~\citep{Bacon17,Florensa17,Vezhnevets17,Osa19} often learn a hierarchical policy given by  $\pi(\vect{a}|\vect{s})=\sum_{o \in \mathcal{O}} \pi(o|\vect{s})\pi(\vect{a}|\vect{s}, o)$, 
where $\vect{s}$, $\vect{a}$ and $o$ denote the state, action, and option, respectively.
Recent studies~\citep{Nachum18,Nachum19,Schaul15} investigated the goal-conditioned policies $\pi(\vect{a}|\vect{s}, \vect{g}) $, where $\vect{g}$ denotes the goal. 
These methods can be interpreted as an approach that models diverse behaviors with a policy conditioned on the latent variable.
However, the problem setting is different from ours because our problem formulation does not involve the Markov decision process.
The details of the difference in problem settings between our and RL studies are discussed in the Appendix~C.

\section{Experiments}
\subsection{Optimization of test objective functions}
To investigate the behavior of LSMO, we applied LSMO to maximize several test functions, which have an infinite set of optimal points.
For visualization, we used the objective functions that take in a two-dimensional input.
The details of the test functions are described in the Appendix~A.2.
The latent variable $\vect{z}$ is one-dimensional in this experiment.
For simplicity, we designed the objective functions such that their maximum value was 1.0.
As a baseline, we evaluated the CEM in which a mixture model of 20 Gaussian distributions was used as the sampling distribution~\citep{Boer05,Kroese06}.
Quantitative results in this section are based on three runs with different random seeds, unless otherwise stated. 

\begin{figure}
	\centering
	\begin{subfigure}[t]{0.2\columnwidth}
		\centering
		\includegraphics[width=\textwidth]{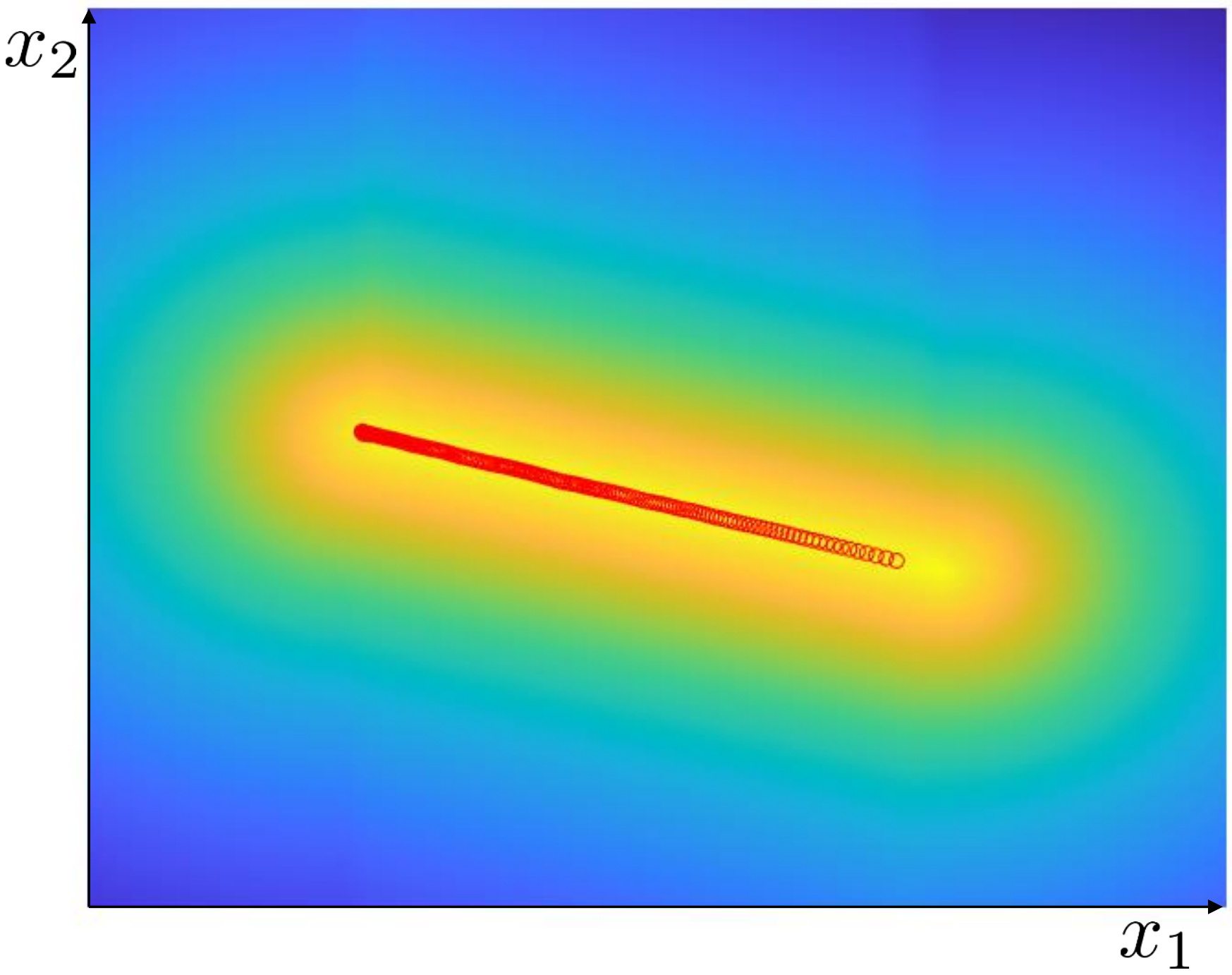}
		\caption{LSMO on Func.~1. }
	\end{subfigure}
	\begin{subfigure}[t]{0.2\columnwidth}
		\centering
		\includegraphics[width=\textwidth]{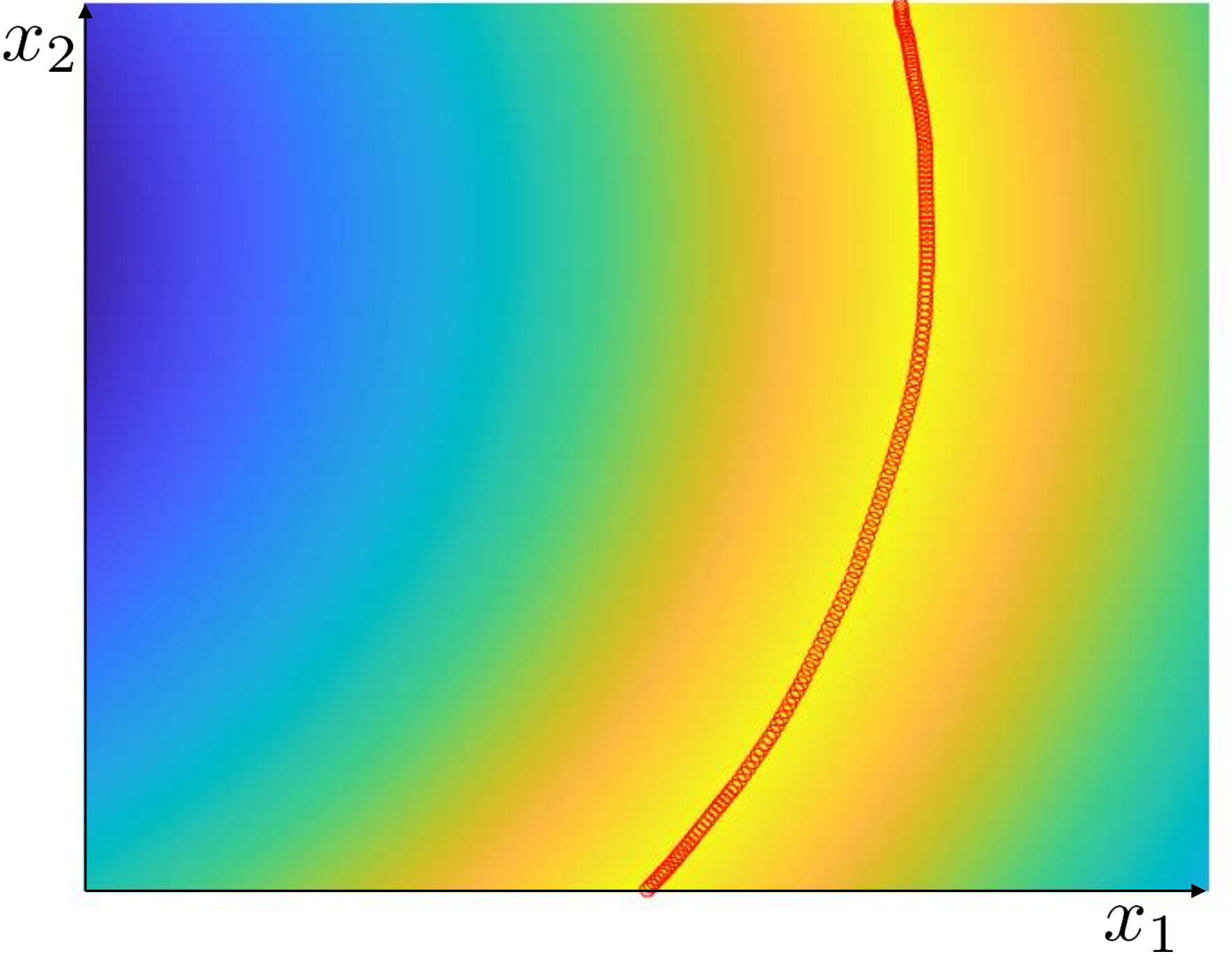}
		\caption{LSMO on Func.~2. }
	\end{subfigure}
	\begin{subfigure}[t]{0.2\columnwidth}
		\centering
		\includegraphics[width=\textwidth]{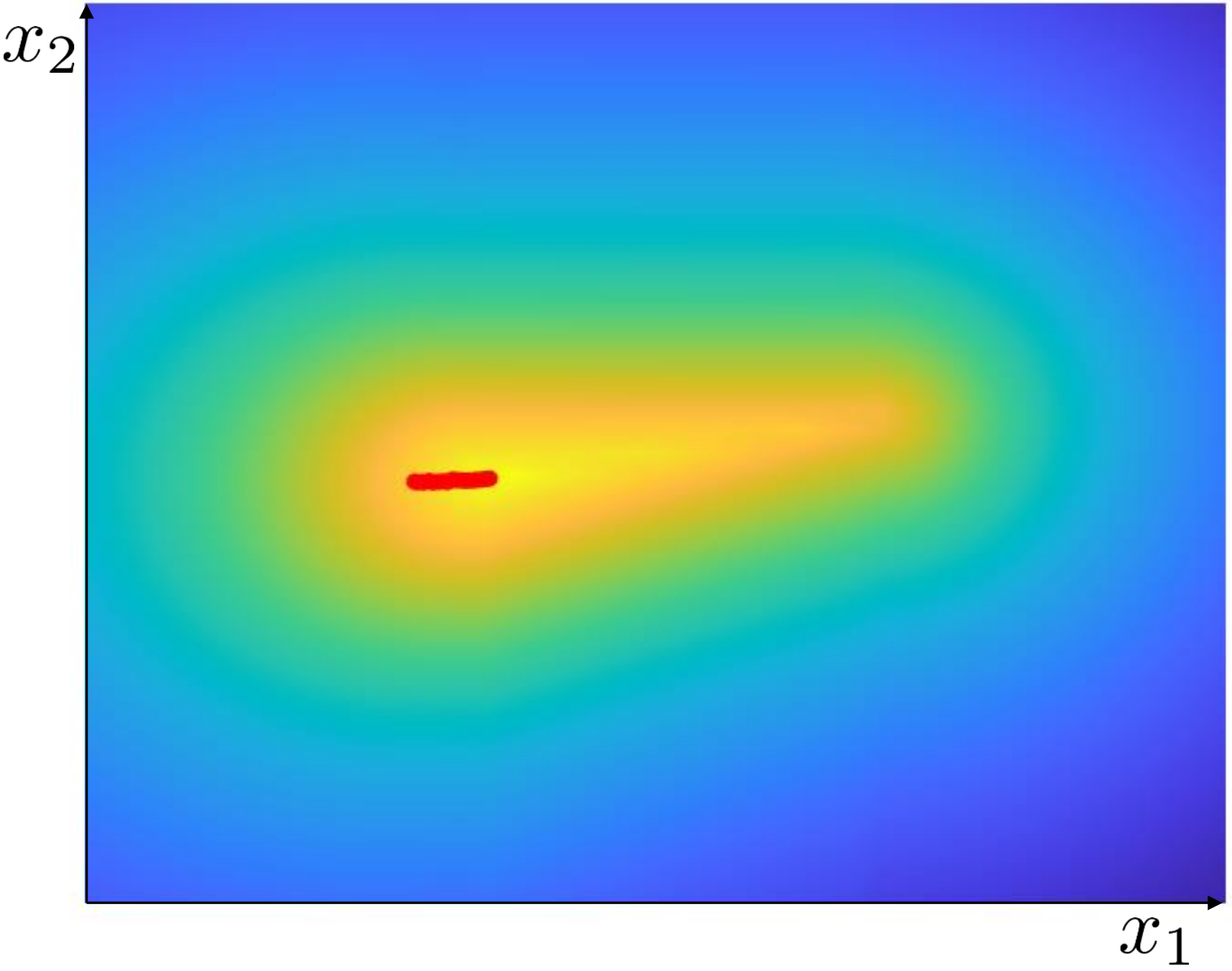}
		\caption{LSMO on Func.~3. }
	\end{subfigure}
	\begin{subfigure}[t]{0.2\columnwidth}
		\centering
		\includegraphics[width=\textwidth]{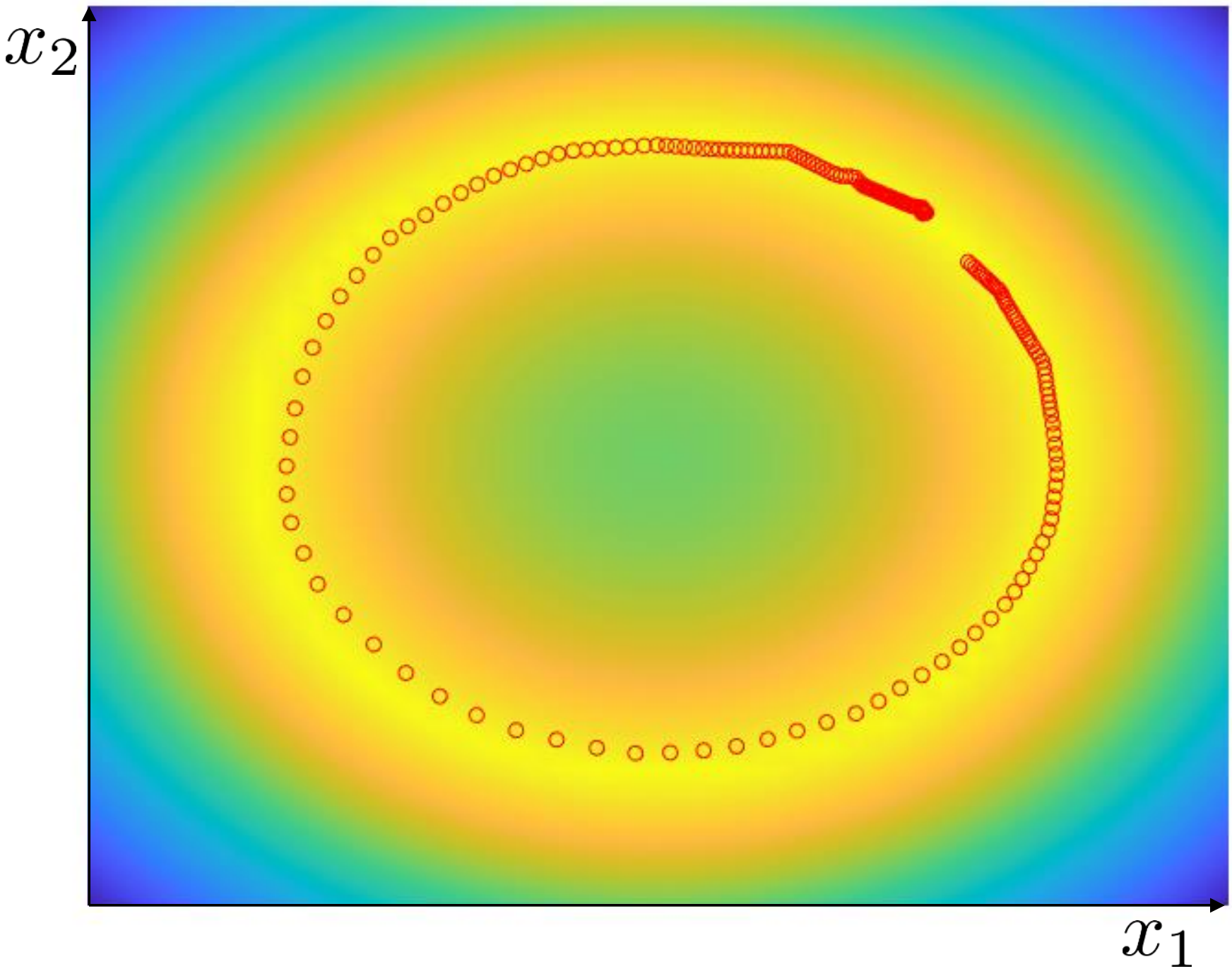}
		\caption{LSMO on Func.~4. }
	\end{subfigure}
	\begin{subfigure}[t]{0.2\columnwidth}
		\centering
		\includegraphics[width=\textwidth]{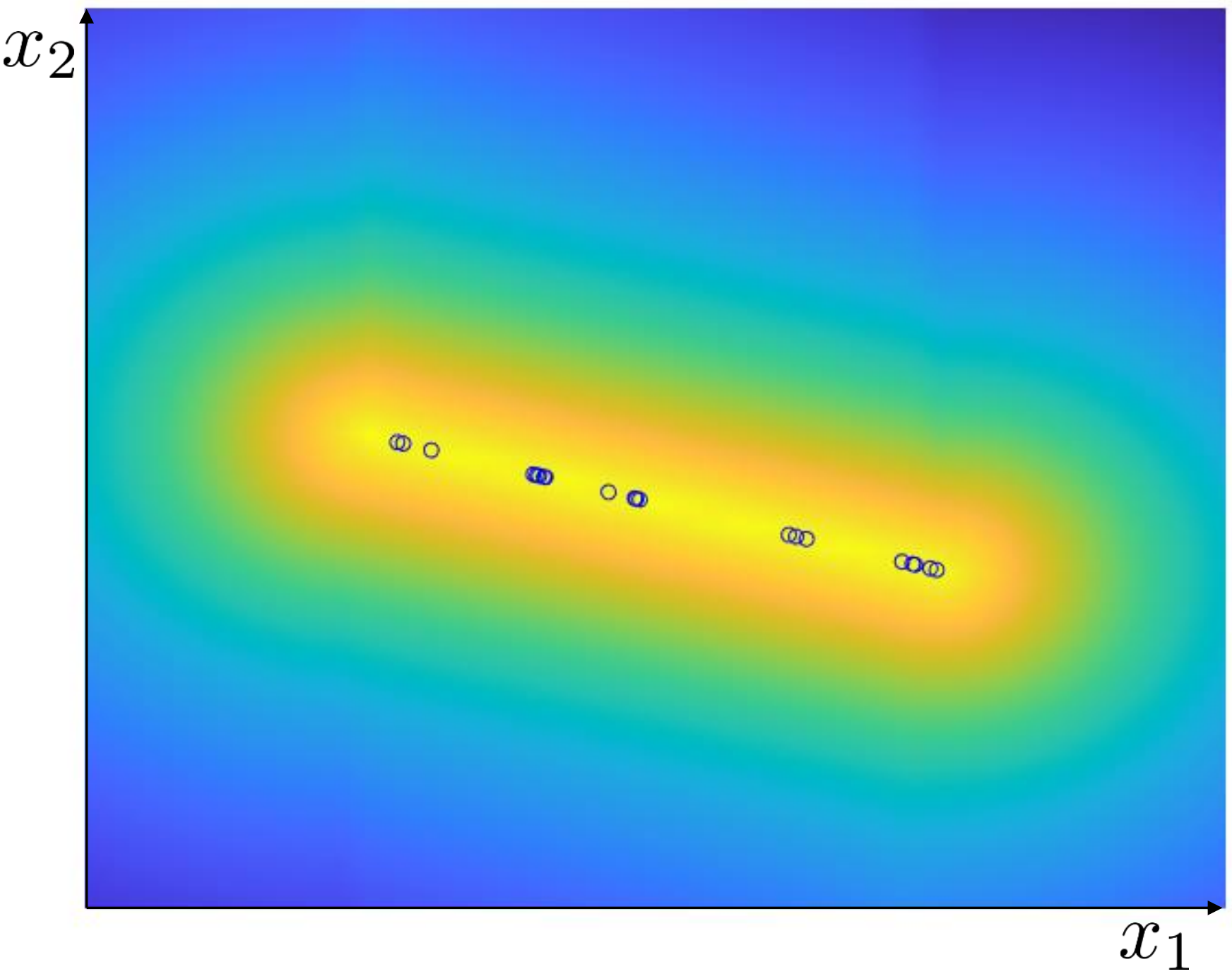}
		\caption{CEM on Func.~1. }
	\end{subfigure}
	\begin{subfigure}[t]{0.2\columnwidth}
		\centering
		\includegraphics[width=\textwidth]{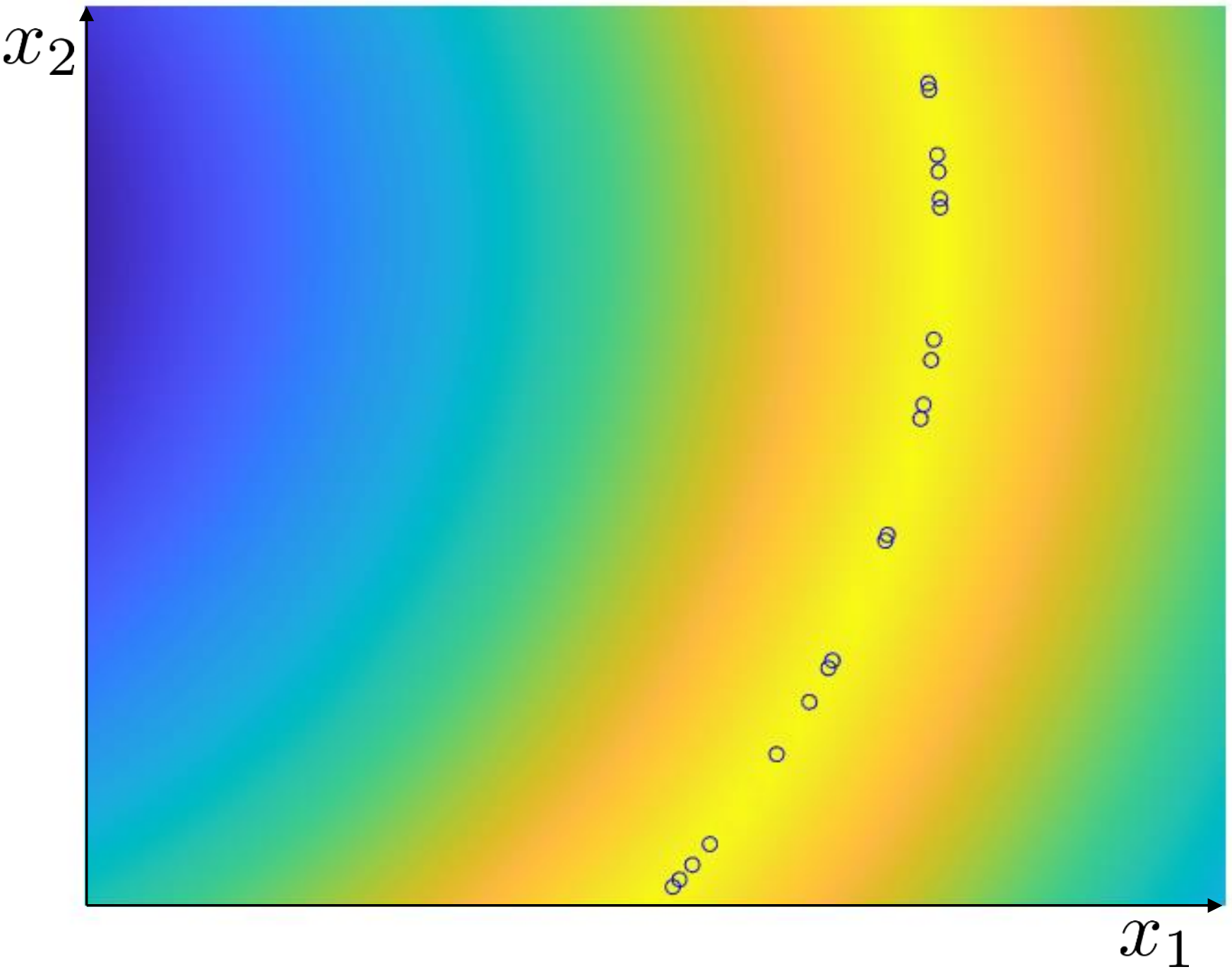}
		\caption{CEM on Func.~3. }
	\end{subfigure}
	\begin{subfigure}[t]{0.2\columnwidth}
		\centering
		\includegraphics[width=\textwidth]{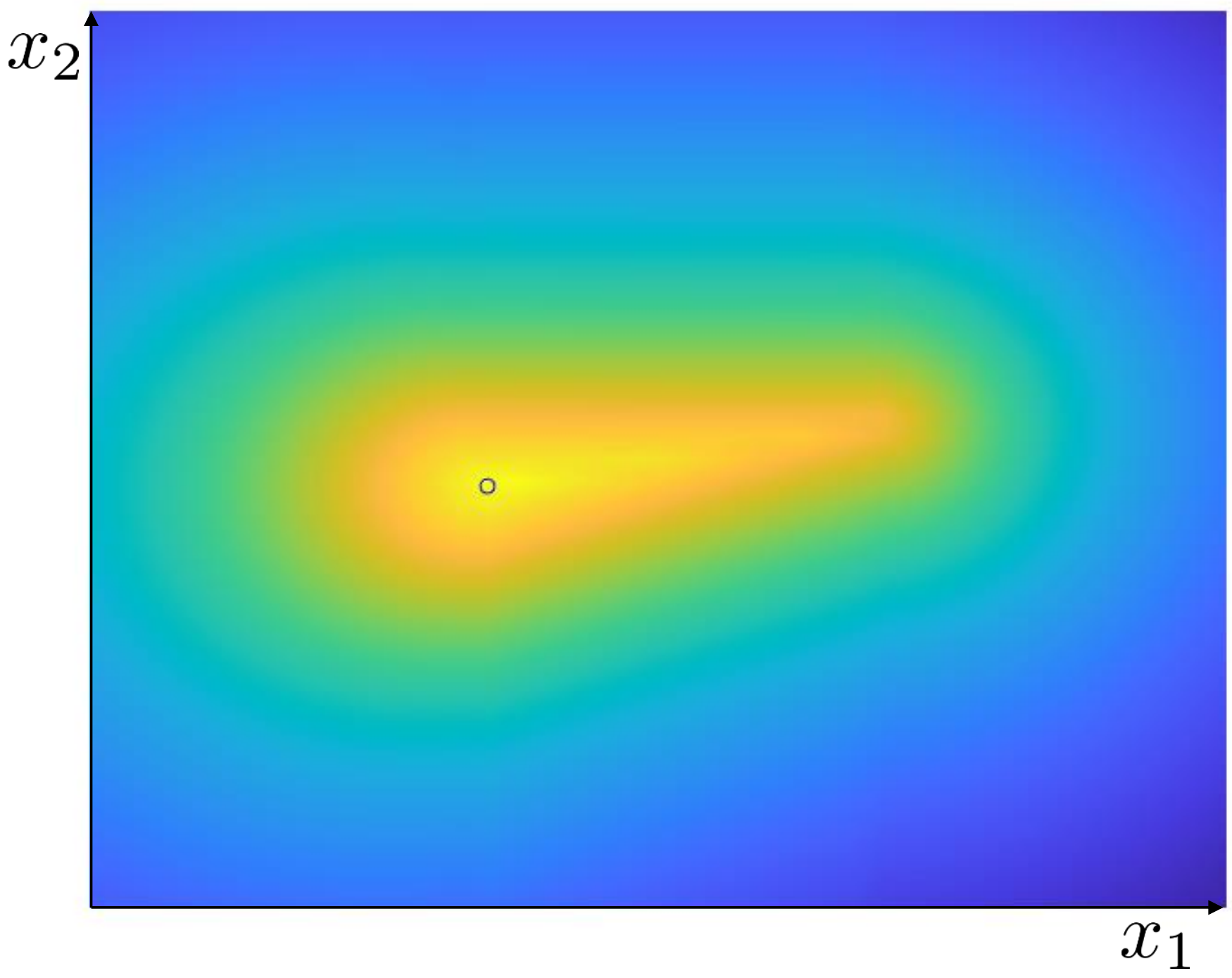}
		\caption{CEM on Func.~3. }
	\end{subfigure}
	\begin{subfigure}[t]{0.2\columnwidth}
		\centering
		\includegraphics[width=\textwidth]{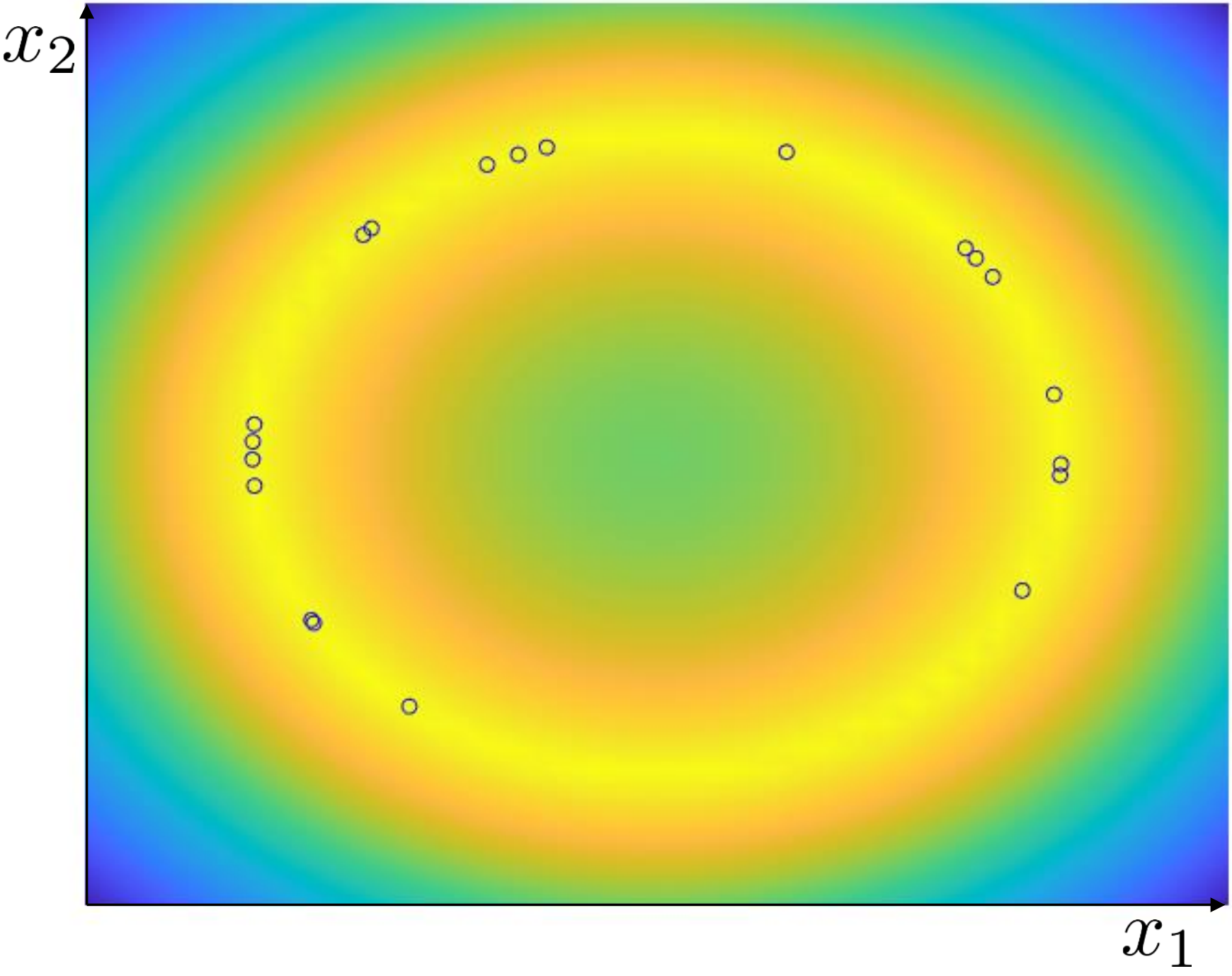}
		\caption{CEM on Func.~4. }
	\end{subfigure}
	\caption{Behavior of LSMO when optimizing the toy objective function. The warmer color represents a higher value of the objective function. In (a)-(d), the outputs of $p_{\vect{\theta}}(\vect{x}| z)$ trained with LSMO are represented by red circles. The outputs of $p_{\vect{\theta}}(\vect{x}|z)$ are generated by linearly changing the value of $z$ in $[-1.28, 1.28]$.  In (e)-(h), the centers of the Gaussian distributions are drawn as blue circles.
	} 
	\label{fig:toy}
\end{figure}

\begin{table}
	\caption{Values of the objective function for the obtained solutions (mean $\pm$ standard deviation)}
	\label{sample-table}
	\centering
	\begin{tabular}{lllll}
		\toprule
		& Func.~1     & Func.~2 & Func.~3 & Func.~4 \\
		\midrule
		LSMO &  $0.996 \pm 9.9\times10^{-3} $  & $0.9996 \pm 4.6\times10^{-4}$ & $0.973\pm0.023$ &  $0.999 \pm  0.0015$   \\
		CEM     &$ 1.00 \pm 3.6\times10^{-6}$       &$ 1.00 \pm 1.3\times10^{-6}$  & $ 1.00 \pm 3.5\times 10^{-6}$ & $ 1.00 \pm 1.3\times 10^{-5}$ \\
		\bottomrule
	\end{tabular}
	\label{tbl:toy}
\end{table}

\paragraph{Qualitative analysis} Outputs of the model $p_{\vect{\theta}}(\vect{x}|\vect{z})$ trained with LSMO and the solutions obtained by CEM are illustrated in Figure~\ref{fig:toy}. 
The results show that the model $p_{\vect{\theta}}(\vect{x}|\vect{z})$ outputs the samples in the region of optimal solutions of the objective function. 
Although CEM also finds multiple solutions for these objective functions, the number of solutions needs to be given and fixed. In addition, the similarity of the obtained solutions is unknown.
On the contrary, the model trained with LSMO can represent an infinite set of solutions. 
Furthermore, the similarity of the value of $\vect{z}$ indicates the similarity of the solutions.
When using LSMO, we can obtain various solutions by changing the value of $\vect{z}$.
However, there is no guarantee that the learned solutions will cover the entire region of optimal points.
For example, although the objective function shown in Figure~\ref{fig:toy}(a) has a set of solutions along a straight line, the manifold learned by LSMO does not cover the entire set of solutions.

\paragraph{Score of solutions} We need to be aware that the output of $p_{\vect{\theta}}(\vect{x}|\vect{z})$ trained by LSMO does not always correspond to the optimal solution. 
The values of the objective function for the solutions obtained with LSMO and CEM are summarized in Table~\ref{tbl:toy}.
Although CEM finds the approximately optimal solutions, the quality of the outputs of  $p_{\vect{\theta}}(\vect{x}|\vect{z})$ trained by LSMO have a larger variance. 
This property can also be observed in Figure~\ref{fig:toy}.
For example, the objective function shown in Figure~\ref{fig:toy}(c) has a unique optimal solution, although the gradient is moderate in one direction.
While CEM finds the optimal point for this objective function, LSMO learns the manifold along the direction in which the gradient of the objective function is moderate.
This result is reasonable because the result of LSMO is inference, rather than exact optimization.  
Therefore, it may be necessary to fine-tune the obtained solution in practice, for example, using a gradient-based method.



\subsection{Motion Planning for a Robotic Manipulator}
To investigate the applicability of LSMO to the optimization of high-dimensional parameters, we evaluated LSMO on motion planning tasks. 
To deal with motion-planning tasks, LSMO is employed in the manner described in Algorithm~\ref{alg:MPLSM}.
We compared the SMTO algorithm, recently proposed in \cite{osa20}. SMTO is an algorithm that finds multiple solutions for motion-planning problems. 
We visualized the planned motions using a simulator based on CoppeliaSim~\citep{coppeliaSim13}. 
Quantitative results in this section are based on three runs with different random seeds, unless otherwise stated. 
Although LSMO is mainly evaluated in simulation, we also verified the applicability to a real robot system. For details of the real robot system, please refer to the Appendix~B.7.

\paragraph{Task setting} The task is to train a model $p_{\vect{\theta}}(\vect{\xi}|\vect{z})$ that generates a collision-free trajectory $\vect{\xi}$ to reach a given goal position from a start position. 
Three task settings are used in this experiment; the details of the task settings are provided in the Appendix~B.4. A trajectory is represented by 50 time steps, and the manipulator has seven DoFs.
Therefore, the dimension of the trajectory parameters  is 350.
For training $p_{\vect{\theta}}(\vect{\xi}|\vect{z})$, we draw 20000 samples from $\beta_{\textrm{traj}}(\vect{\xi})$.
The details of the cost function $\mathcal{C}(\vect{\xi})$ and shaping function $f(\cdot)$ used in this experiment are described in the Appendix~B.1.

\begin{figure}
	\begin{subfigure}[t]{\columnwidth}
		\includegraphics[width=\textwidth]{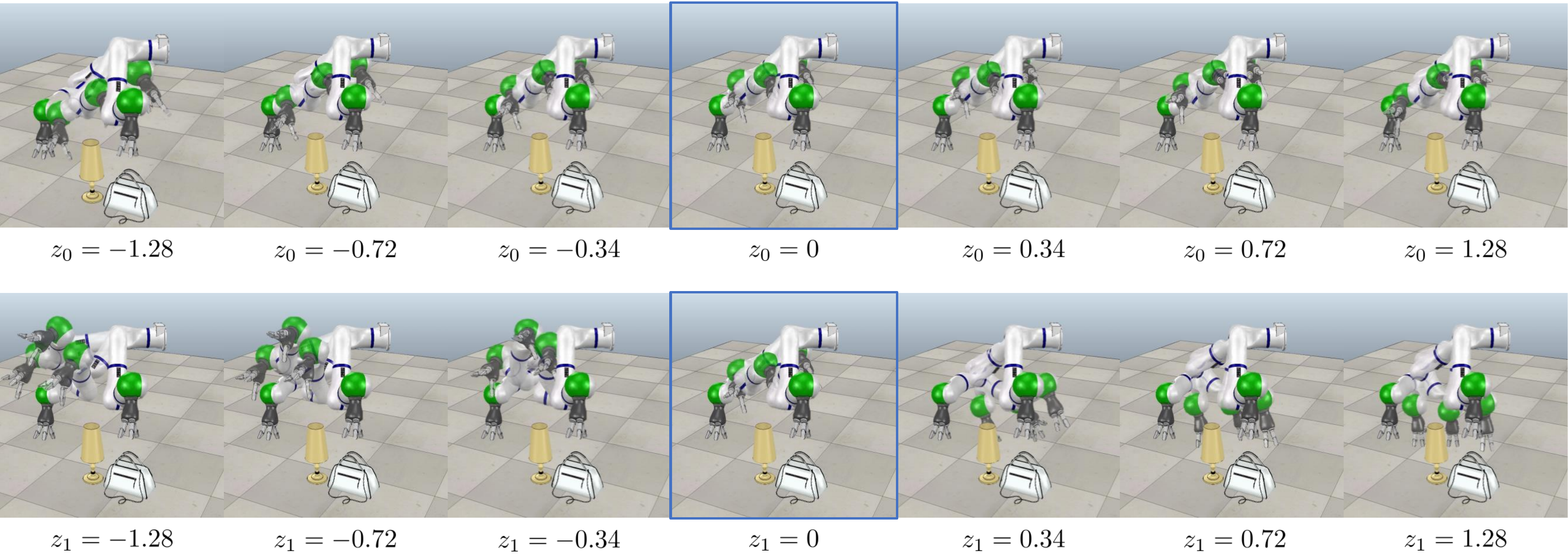}
		\caption{Variation of solutions obtained by LSMO for Task~1. The latent variable $\vect{z}$ is two-dimensional as $\vect{z} = [z_0, z_1]$ in this result. The top row shows the variation with different $z_0$ and the second row shows the variation with different $z_0$. Trajectories indicated by the blue square show the trajectory generated with $z_0=z_1 =0$.} 
	\end{subfigure}
	\begin{subfigure}[t]{\columnwidth}
		\includegraphics[width=\textwidth]{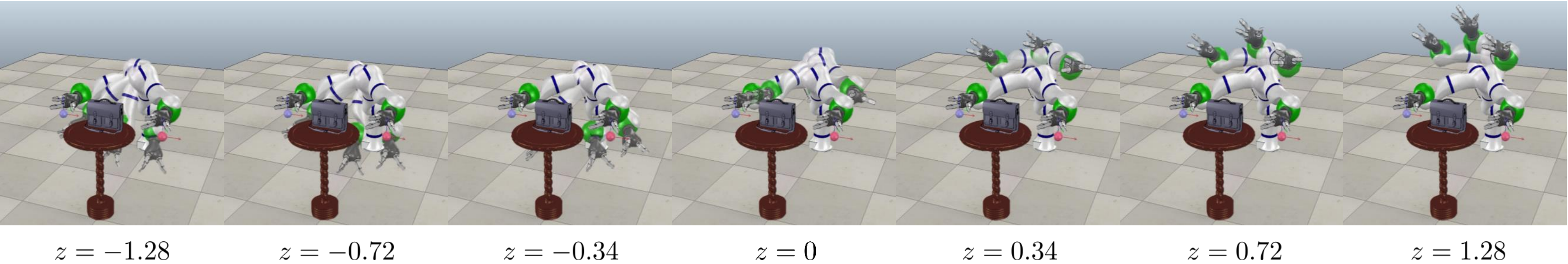}
		\caption{Variation of solutions obtained by LSMO for Task~2. The latent variable $\vect{z}$ is one-dimensional in this result. } 
	\end{subfigure}
	\begin{subfigure}[t]{\columnwidth}
		\includegraphics[width=\textwidth]{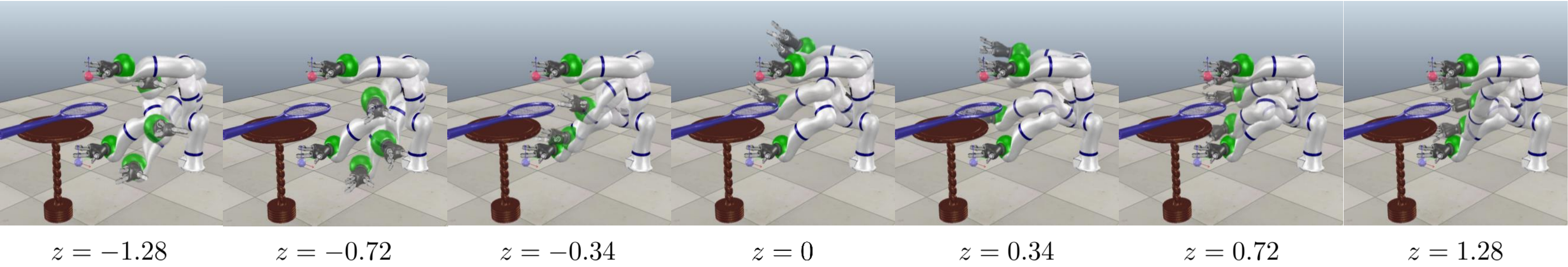}
		\caption{Variation of solutions obtained by LSMO for Task~3. The latent variable $\vect{z}$ is one-dimensional in this result. } 
	\end{subfigure}
	\begin{subfigure}[t]{0.39\columnwidth}
		\includegraphics[width=\textwidth]{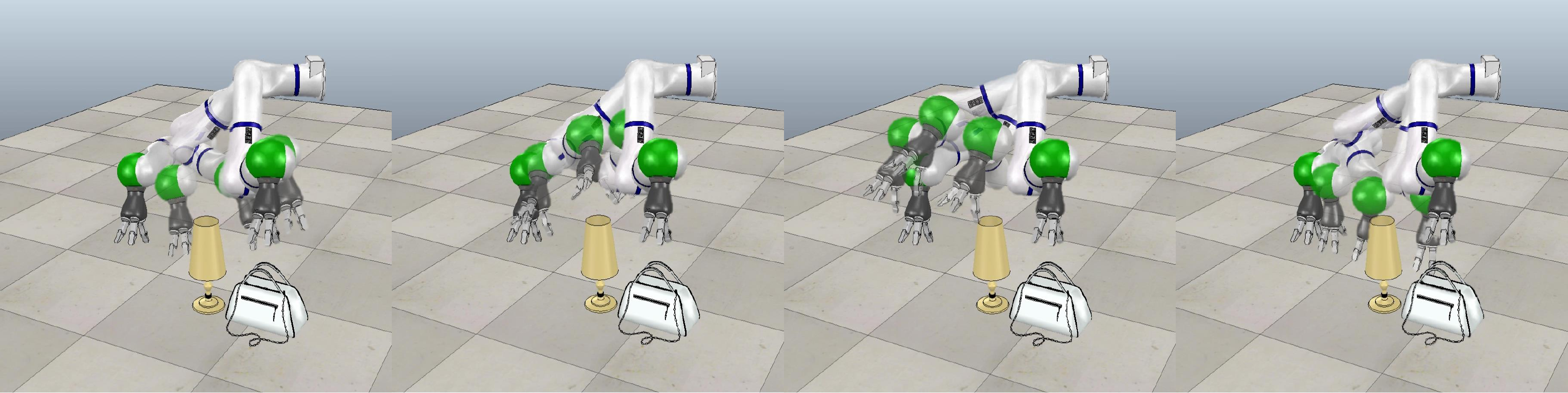}
		\caption{SMTO on Task~1. } 
	\end{subfigure}
	\begin{subfigure}[t]{0.39\columnwidth}
		\includegraphics[width=\textwidth]{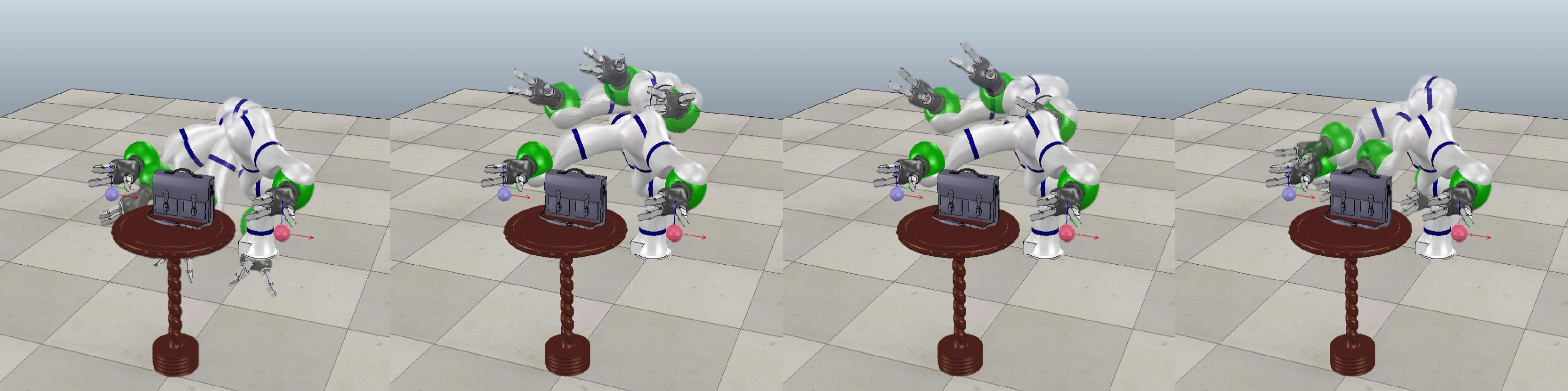}
		\caption{SMTO on Task~2. } 
	\end{subfigure}
	\begin{subfigure}[t]{0.19\columnwidth}
		\includegraphics[width=\textwidth]{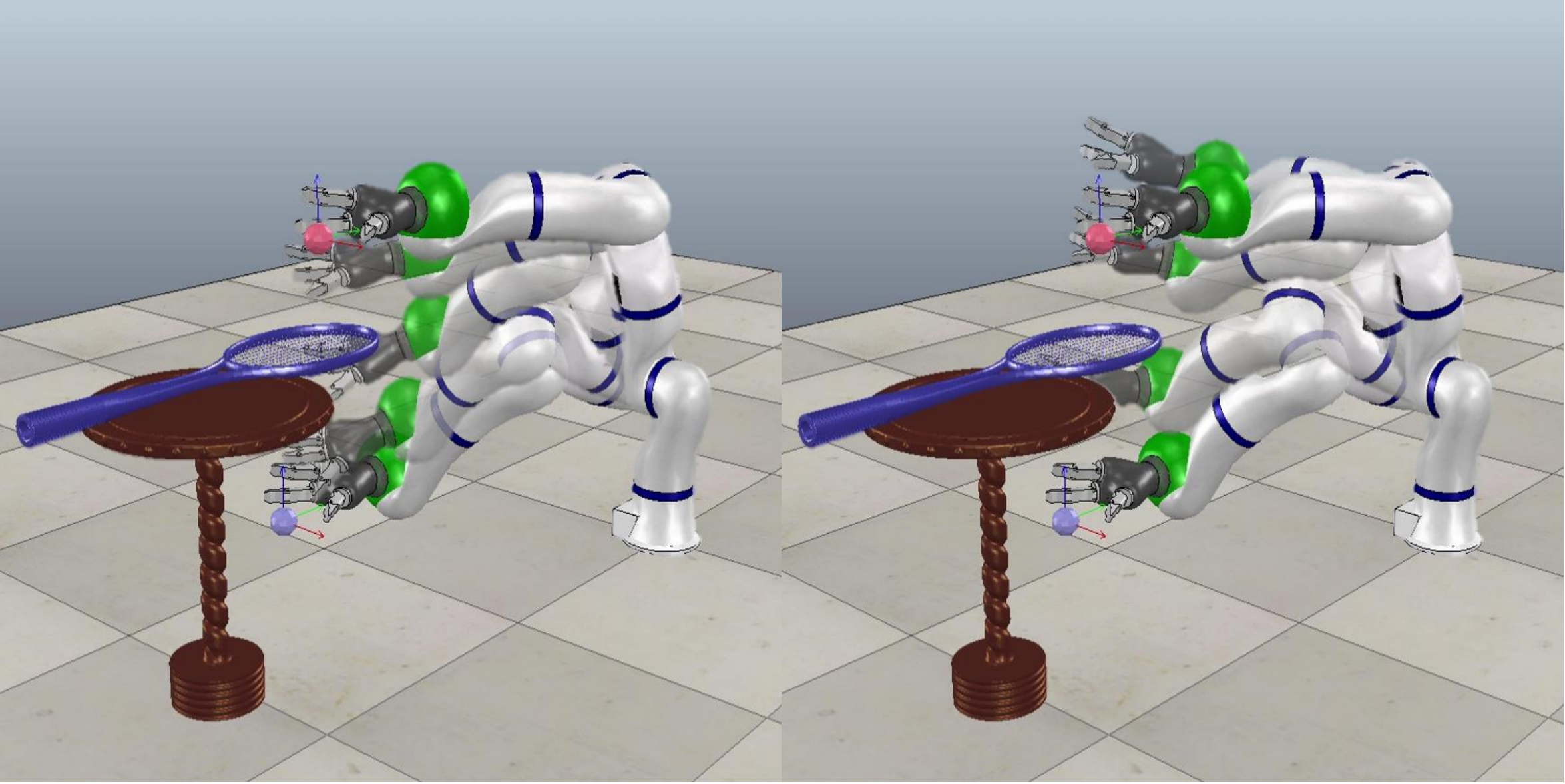}
		\caption{SMTO on Task~3. } 
	\end{subfigure}
	\caption{Qualitative comparison of LSMO and SMTO.
		(a)-(c): Solutions generated by the model $p_{\vect{\theta}}(\vect{\xi}|\vect{z})$ trained with LSMO with different values of $z$. The model $p_{\vect{\theta}}(\vect{\xi}|\vect{z})$ represents an infinite set of solutions, and solutions can be continuously varied by changing the value of $z$.
		(d)-(f): Solutions obtained with SMTO. SMTO is limited to generating a finite number of solutions.}
	\label{fig:LSMO_mp}
	\vspace{-.2cm}
\end{figure}

%

\paragraph{Qualitative comparison} 
The outputs of the model $p_{\vect{\theta}}(\vect{\xi}|\vect{z})$ trained with LSMO, which were not fine-tuned, are depicted in Figure~\ref{fig:LSMO_mp}(a)-(c)
The model $p_{\vect{\theta}}(\vect{\xi}|\vect{z})$ generates various collision-free trajectories with given start and goal configurations.
This result indicates that LSMO is applicable to the optimization of high-dimensional parameters.
In addition, it is evident that a trajectory generated by $p_{\vect{\theta}}(\vect{\xi}|\vect{z})$ can be continuously deformed into another, which indicates that the model trained with LSMO represents an infinite set of homotopic solutions. 
Figure~\ref{fig:LSMO_mp}~(a) also shows that, when learning the latent variable is two-dimensional, each dimension of the latent variable encodes a different type of variation on Task~1. 
However, the appropriate dimensionality of the latent variable depends on the tasks.
A comparison of the performance when using the two-dimensional and one-dimensional latent variables is provided in the Appendix~B.5.
The number of solutions found by SMTO was 2-4 on the three tasks, as shown in Figure~\ref{fig:LSMO_mp}(d)-(f).
Although SMTO finds multiple solutions, the number of solutions are limited and the similarity of the solutions is not indicated.
On the contrary, $p_{\vect{\theta}}(\vect{\xi}|\vect{z})$ trained with LSMO can generate various solutions by changing the value of $\vect{z}$.
Since the value of $\vect{z}$ indicates the similarity of solutions, it is easy to generate a ``middle'' solution between solutions.

\begin{table}[b]
	\vspace{-0.5cm}
	\caption{Values of the objective function for motion planning (higher is better.)}
	\centering
	\begin{tabular}{llll}
		\toprule
		& Task~1     & Task~2  & Task~3	 \\
		\midrule
		LSMO w/o fine-tuning & $-2.95 \pm 1.33 $  & $ -2.24 \pm 1.15$ & $-3.24 \pm 0.92 $ \\
		LSMO w/ fine-tuning &  $-1.89 \pm 0.31 $  & $ - 1.66 \pm 0.15$ & $-2.02 \pm 0.25 $\\
		SMTO     &$ - 1.64 \pm 0.09$ &$ - 1.53 \pm 0.176$  & $-1.89 \pm 0.04 $\\
		CHOMP w/ random init.     &$ - 1.82 \pm 0.26$ &$ - 1.57 \pm 0.39$ &  $-2.06 \pm 0.41 $ \\
		\bottomrule
	\end{tabular}
	\vspace{-0.7cm}
	\label{tbl:motion}
\end{table}


\subsection{Computation time}
We show the computation time for generating a trajectory on motion planning tasks in Table~\ref{tbl:motion_time}.
We used a laptop with Intel\textsuperscript{\textregistered} Core-i7-8750H CPU and NVIDIA Geforce\textsuperscript{\textregistered} GTX 1070 GPU for the experiment.
Although LSMO can generate a solution quickly after training the model $p_{\vect{\theta}}(\vect{\xi}|z)$, the time required for training $p_{\vect{\theta}}(\vect{\xi}|z)$ takes approximately 30~min for each task.
The computation time for SMTO is the time required for generating four solutions as shown  Figure~4.
Although CHOMP generates only a single solution, we show the computation time as a baseline. 
To evaluate the computation time for CHOMP, we initialized the trajectory by sampling from the proposal distribution~$\beta_{\textrm{traj}}(\vect{\xi})$ in (15), and computed the average and the standard deviation from ten runs.
Table~\ref{tbl:motion_time} shows that the time required for fine-tuning of LSMO with CHOMP is substantially shorter than that for the motion planning with CHOMP starting from randomized initialization. 
This result indicates that the model trained with LSMO outputs a trajectory close to the optimal solutions.
\begin{table}
	\caption{Computation time in motion planning tasks. }
	\centering
	\begin{tabular}{llll}
		\toprule
		& Task~1     & Task~2  & Task~3	 \\
		\midrule
		Fine-tuning of LSMO with CHOMP &  $0.78 \pm 0.51 $~[sec]  & $ 0.34 \pm 0.20$~[sec] & $0.93 \pm 0.21$~[sec]  \\
		\midrule
		SMTO     &$ 52.3\pm 0.37$~[sec] &$84.1\pm 29.5$~[sec]  & $85.7.3\pm 29.3$~[sec] \\
		CHOMP with randomized initialization     &$6.5\pm 14.2$~[sec] &$ 1.4 \pm 0.76$~[sec] & $6.6 \pm 13.6$~[sec]  \\
		\bottomrule
	\end{tabular}
	\vspace{-0.3cm}
	\label{tbl:motion_time}
\end{table}

\paragraph{Scores of solutions} The scores of the obtained solutions are summarized in Table~\ref{tbl:motion}.
As a baseline, we also show the results when we run CHOMP with random initialization, although CHOMP can find only one local optima for each run. The reported scores of CHOMP are the average and standard deviation of ten runs.
Although the values of the solutions obtained with SMTO are better than those obtained with LSMO,
all the solutions obtained by LSMO without fine-tuning are collision-free and executable in a real robot.
As generating a trajectory with the model $p_{\vect{\theta}}(\vect{\xi}|\vect{z})$ is inference rather than optimization, 
it is natural that their scores are lower than the results of optimization-based methods.
However, with quick fine-tuning, we can obtain solutions with comparable scores.

\paragraph{Limitations of LSMO} As reported in this section, a limitation of LSMO is that it requires fine-tuning to obtain the exact solution. However, we believe that this limitation is not difficult to overcome because the output of $p_{\vect{\theta}}(\vect{\xi}|\vect{z})$ is reasonably close to an optimal point and can be quickly fine-tuned.  
Another limitation is that training $p_{\vect{\theta}}(\vect{\xi}|\vect{z})$ is time-consuming compared with existing motion planners.
However, once we train the model $p_{\vect{\theta}}(\vect{\xi}|\vect{z})$, we can generate various solutions substantially faster than modifying the objective function and re-running a standard motion planning method.
The time-consuming process of training $p_{\vect{\theta}}(\vect{\xi}|\vect{z})$ with LSMO is drawing and evaluating samples, which are used for computing the loss function in \eqref{eq:loss}.
Although our implementation is not optimized, this sampling process can be parallelized for fast computation.


\subsection{Applicability to a real robotic system}
To verify the applicability of LSMO to a real robot system, we applied LSMO to motion planning for a real robotic system. 
In this experiment, we used COBOTTA produced by Denso Wave Inc., which has six DoFs.
The network architecture was the same as in Table~\ref{tbl:param_mp}.
After generating trajectories with 50 time steps using $p_{\vect{\theta}}(\vect{\xi}|\vect{z})$, the trajectories are interpolated with cubic spline in order to send it to the robotic system.
In this experiment, the time for training $p_{\vect{\theta}}(\vect{\xi}|\vect{z})$ is about 30 min using the laptop used in the previous experiments.

\begin{figure}[tb]
	\includegraphics[width=\textwidth]{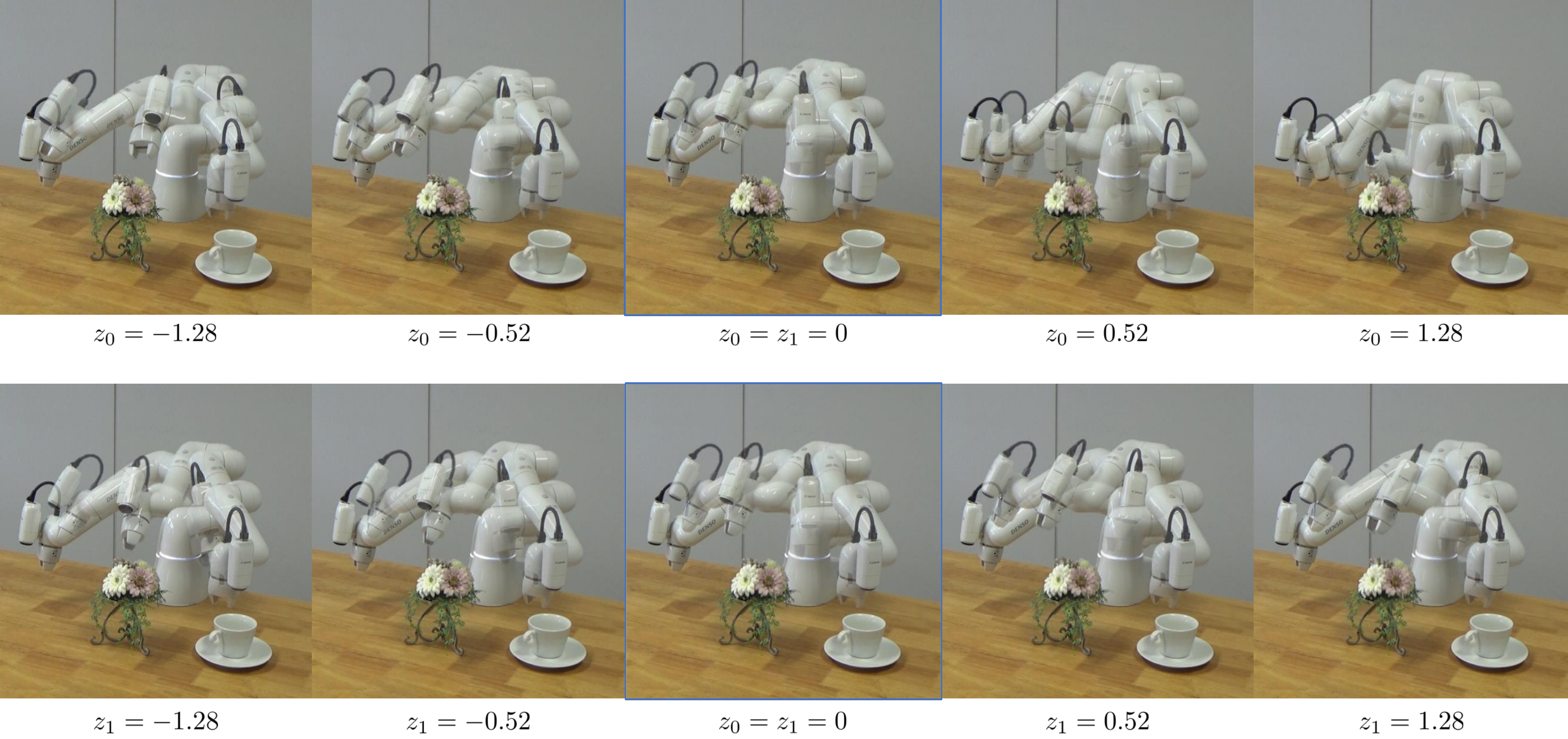}
	\caption{Visualization of solution obtained by LSMO for the task with a real robot. The latent variable $\vect{z}$ is two-dimensional as $\vect{z} = [z_0, z_1]$ in this result. The top row shows the variation with different $z_0$ and the second row shows the variation with different $z_1$. Trajectories indicated by the blue square show the trajectory generated with $z_0=z_1 =0$.}
	\label{fig:cobotta}
\end{figure}

The task is to plan a trajectory to reach the goal configuration, while avoiding an obstacle.
We show the results with the two-dimensional latent variable in Figure~\ref{fig:cobotta}.
In the case of the two-dimensional latent variable, each dimension of the latent variable encodes different variations of solutions; the channel $z_0$ encodes the height of the end-effector, and the channel $z_1$ encodes the orientation of the end-effector.
The end-effector goes over the obstacle when $z_0=-1.28$, while the end-effector goes behind the obstacle when $z_0=1.28$.
When varying $z_1$, the orientation of the end-effector changes, although the height of the end-effector does not change significantly.

\section{Discussion and Future Work}
We presented LSMO, which is an algorithm for learning the manifold of solutions in optimization.
Our contribution is to shed light on the approach of training a model that represents an infinite set of solutions, while existing methods are often limited to generating a finite set of solutions.
In our experiments, we show that a set of homotopic solutions can be obtained with LSMO on motion planning tasks, which involve hundreds of parameters.
We believe that a framework for learning the solution manifold gives users an intuitive way to go through various solutions by changing the value of the latent variables.
Our work focuses on learning the continuous latent variable in optimization. 
However, when separate sets of solutions exist, it will be necessary to learn the discrete latent variable in addition to the continuous latent variable.
In such cases, learning the discrete latent variable in LSMO should be possible with the Gumbel-Softmax trick~\cite{Jang17,Maddison17}. We will investigate this direction in future work.

\appendix

\section{Details of Toy Function Optimization Tasks}

\subsection{Shaping function}
Given a set of samples drawn from a proposal distribution $\mathcal{D}=\{ \vect{x}_i \}^N_{i=1}$, we used the shaping function given by
\begin{align}
f(R(\vect{x})) = \exp\left( \frac{10\big( R(\vect{x}) - R_{\textrm{max}} \big) }{ R_{\textrm{max}} - R_{\textrm{min}} } \right),
\end{align}
where $R_{\textrm{max}} = \max_{\vect{x} \in \mathcal{D}} R(\vect{x})$ and $R_{\textrm{min}} = \min_{\vect{x} \in \mathcal{D}} R(\vect{x})$.

\subsection{Toy functions}
The definitions of the toy functions used in the experiment are given as follows.
The figures plot the range $x_1 \in [0, 2]$ and  $x_2 \in [0, 2]$.
The toy function~1 is given by
\begin{align}
R(x_1, x_2) = \exp( - d )
\end{align}
where
\begin{align}
d = 
\left\{
\begin{array}{cll}
((x_2 - 1.05)^2 + (x_1 - 0.5)^2)^{0.5}, & \textrm{if} & x_1 < 0.5, \\
\frac{| -0.3 x_1 - x_2 + 1.2 |}{( 0.09 + 1 )^2}   , & \textrm{if} & 0.5 < x_1 < 1.5, \\
((x_2 - 0.75)^2 + (x_1 - 1.5)^2)^{0.5},& \textrm{if} & x_1 \geq 1.5.
\end{array}
\right. 
\end{align}
%
%

The toy function~2 is given by
\begin{align}
R(x_1, x_2) = \exp( - d /10 )
\end{align}
where
\begin{align}
d = | (x_2 - 1.5)^2 + (x_1 + 1)^2 - 2.5 |.
\end{align}
%
%

The toy function~3 is given by
\begin{align}
R(x_1, x_2) = \exp\big( - (d + 0.2x_2 + 0.14) \big)
\end{align}
where
\begin{align}
d = 
\left\{
\begin{array}{cll}
((x_2 - 0.94)^2 + (x_1 - 0.7)^2)^{0.5}, & \textrm{if} & x_1 < 0.7, \\
\frac{| 0.2 x_1 - x_2 + 0.8 |}{( 0.04 + 1 )^2}   , & \textrm{if} & 0.7 < x_1 < 1.4, \\
((x_2 - 1.08)^2 + (x_1 - 1.4)^2)^{0.5},& \textrm{if} & x_1 \geq 1.4.
\end{array}
\right. 
\end{align}
%
%

The toy function~4 is given by
\begin{align}
R(x_1, x_2) = \exp( - d /10 )
\end{align}
where
\begin{align}
d = | (x_2 - 1)^2 + (x_1 - 1)^2 - 0.5 |.
\end{align}
%
%

\subsection{Proposal distribution}
\begin{figure}
	\centering
	\begin{subfigure}[t]{0.245\columnwidth}
		\centering
		\includegraphics[width=\textwidth]{RewardSingle2}
		\caption{Func.~1. }
	\end{subfigure}
	\begin{subfigure}[t]{0.245\columnwidth}
		\centering
		\includegraphics[width=\textwidth]{RewardCurve2}
		\caption{Func.~2. }
	\end{subfigure}
	\begin{subfigure}[t]{0.245\columnwidth}
		\centering
		\includegraphics[width=\textwidth]{RewardSingleSlope}
		\caption{Func.~3. }
	\end{subfigure}
	\begin{subfigure}[t]{0.245\columnwidth}
		\centering
		\includegraphics[width=\textwidth]{RewardCircle}
		\caption{Func.~4. }
	\end{subfigure}
	\begin{subfigure}[t]{0.245\columnwidth}
		\centering
		\includegraphics[width=\textwidth]{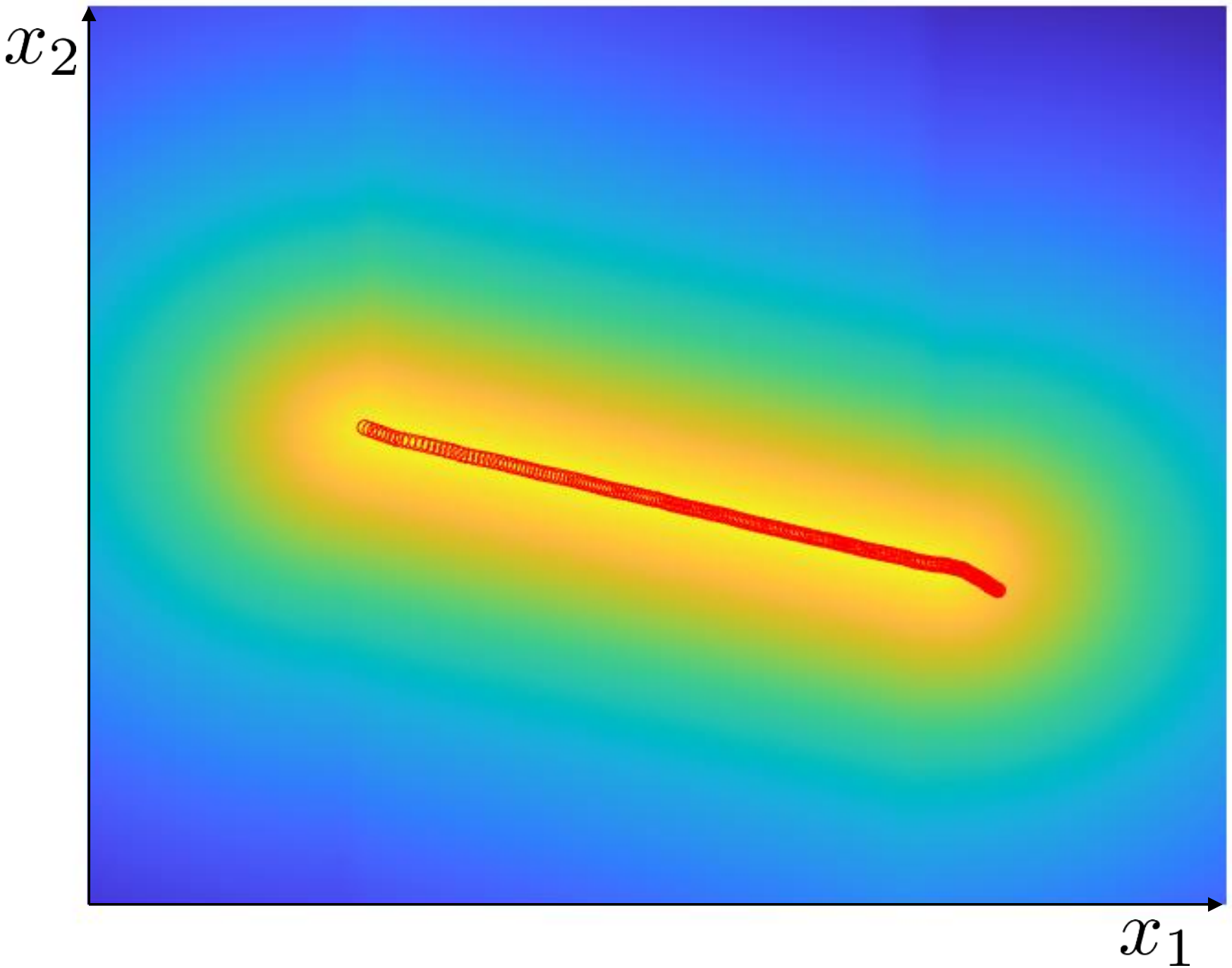}
		\caption{Func.~1. }
	\end{subfigure}
	\begin{subfigure}[t]{0.245\columnwidth}
		\centering
		\includegraphics[width=\textwidth]{RewardSingle2_norm}
		\caption{Func.~3. }
	\end{subfigure}
	\begin{subfigure}[t]{0.245\columnwidth}
		\centering
		\includegraphics[width=\textwidth]{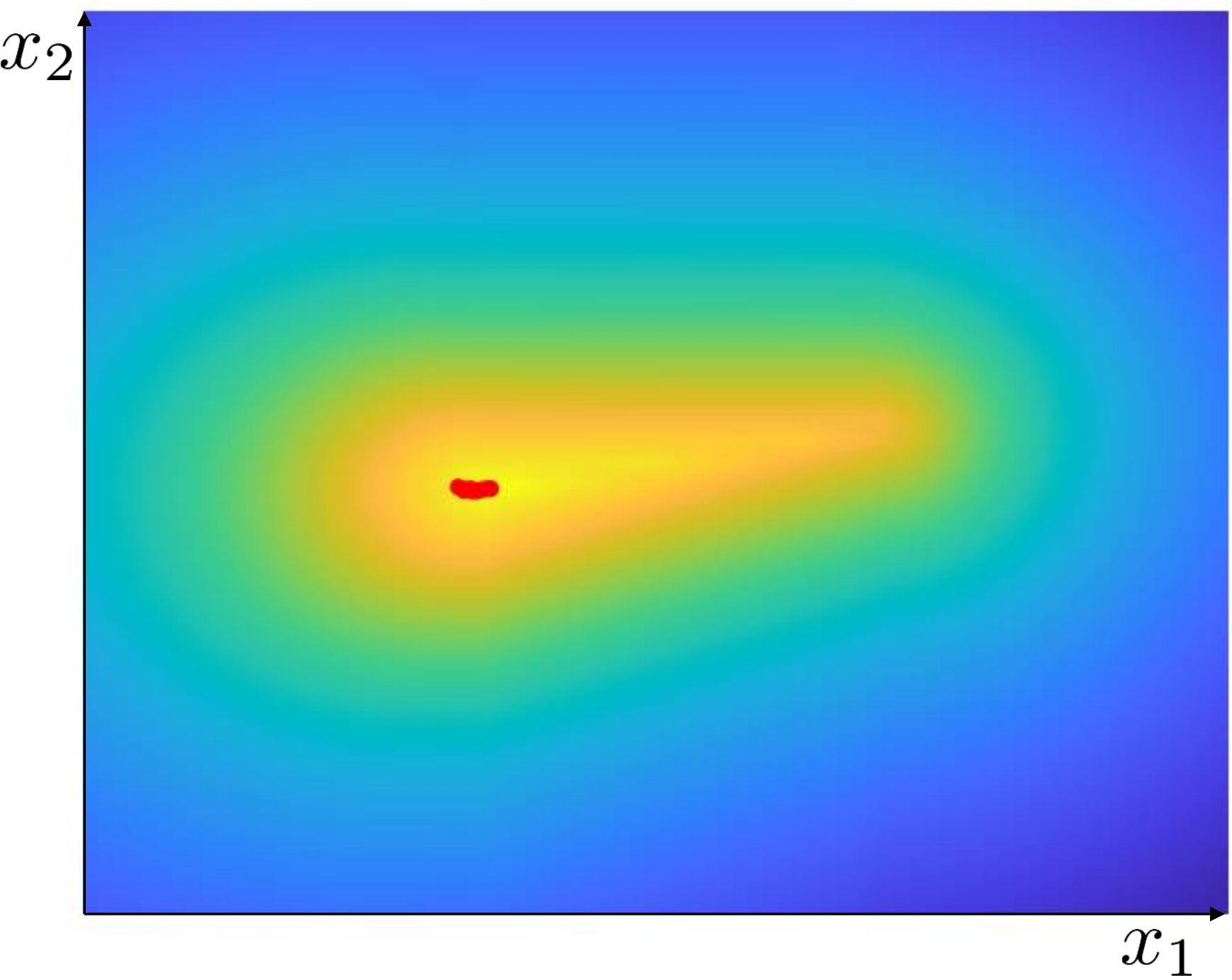}
		\caption{Func.~3. }
	\end{subfigure}
	\begin{subfigure}[t]{0.245\columnwidth}
		\centering
		\includegraphics[width=\textwidth]{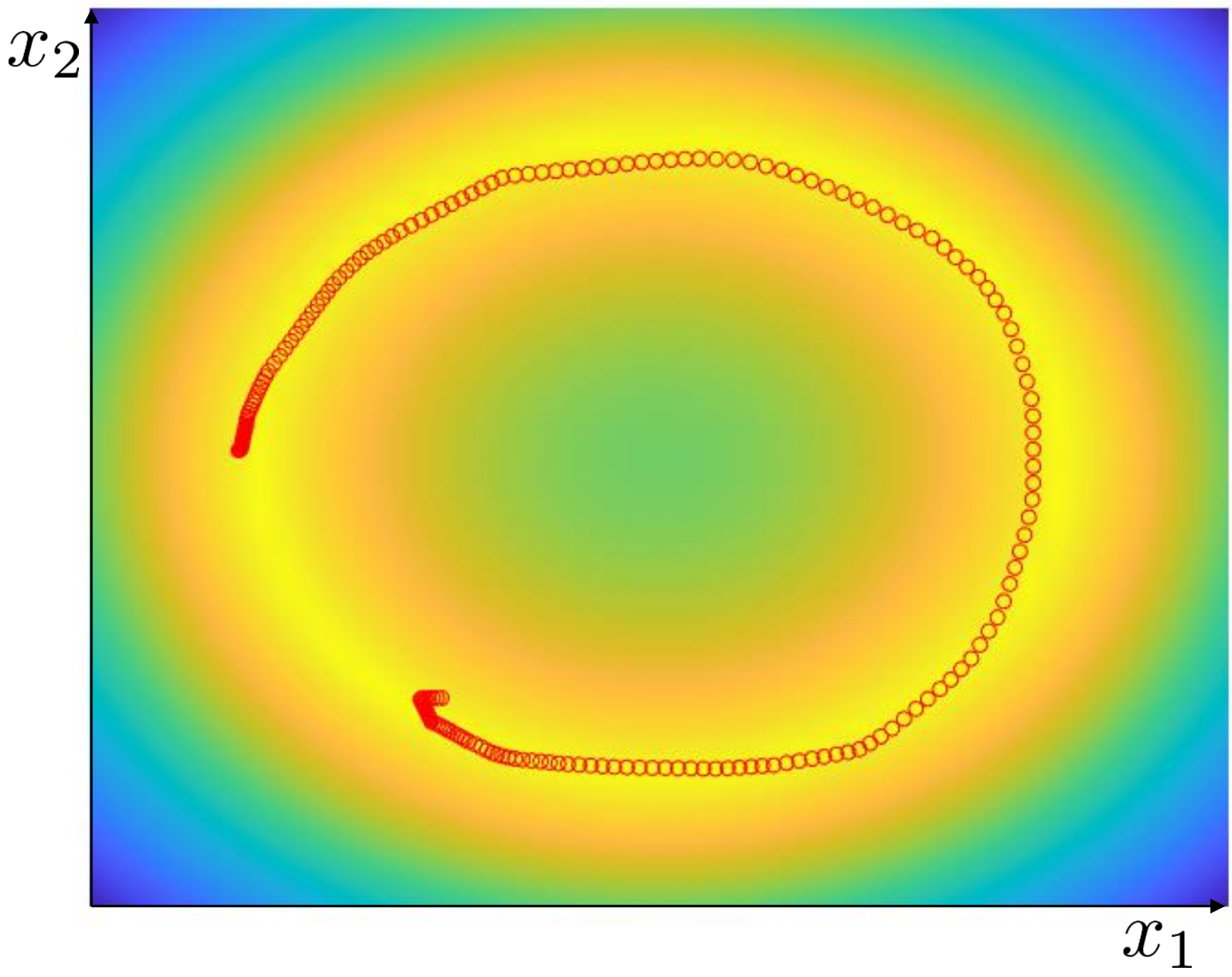}
		\caption{Func.~4. }
	\end{subfigure}
	\caption{Behavior of LSMO when optimizing the toy objective function. The warmer color represents a higher value of the objective function. The outputs of $p_{\vect{\theta}}(\vect{x}| z)$ trained with LSMO are shown as red circles. (a)-(d): Results using the uniform distribution as the initial proposal distribution. (e)-(h): Results using the normal distribution as the initial proposal distribution.
	} 
	\label{fig:toy_prop}
\end{figure}

In the result shown in Figure~3 in the main manuscript, we used a uniform distribution as the initial proposal distribution. 
In the second iteration, we used samples given by
\begin{align}
\tilde{\vect{x}} = \vect{x} + \vect{\epsilon}, \ \textrm{where} \ \vect{x} \sim p_{\vect{\theta}}(\vect{x}) = \int p_{\vect{\theta}}(\vect{x}|\vect{z})p(\vect{z}) \intd \vect{z}, \ \vect{\epsilon} \sim U(-0.1, 0.1).
\end{align}
To analyze the effect of the proposal distribution, we show the results using a normal distribution, $x_i \sim \mathcal{N}(1.0, 0.5)$ for $i=1, 2$, as the initial distribution. The values of the outputs of $ p_{\vect{\theta}}(\vect{x}|\vect{z})$ are summarized in Table~\ref{tbl:toy_prop} and Figure~\ref{fig:toy_prop}.
When we evaluate the scores in Table~1 in the main manuscript and Table~\ref{tbl:toy_prop} in this supplementary, we linearly interpolate $z$ in $[-1.28, 1.28]$ and generate 200 samples from $p_{\vect{\theta}}(\vect{x}|\vect{z})$. These values are chosen because $P(z<-1.28)=0.1$ and $P(z<1.28)=0.9$ when $p(z) = \mathcal{N}(0, 1)$. 

As in other importance sampling methods, the quality of the results is dependent on the choice of the proposal distribution.
For the optimization of the toy function~3, the use of the normal distribution resulted in better performance than did the uniform distribution.
However, in other functions, the use of a uniform distribution outperformed the results of the normal distribution.

\begin{table}
	\caption{Values of the objective function for the obtained solutions (mean $\pm$ standard deviation.)}
	\centering
	\begin{tabular}{lllll}
		\toprule
		& Func.~1     & Func.~2 & Func.~3 & Func.~4 \\
		\midrule
		uniform &  $0.996 \pm 9.9\times10^{-3} $  & $0.9996 \pm 4.6\times10^{-4}$ & $0.973\pm0.023$ &  $0.999 \pm  0.0015$    \\
		normal     &$ 0.989 \pm 0.023$       &$ 0.9992 \pm 6.6\times10^{-4}$  & $ 0.928 \pm 0.0061$ & $ 0.927 \pm 0.004$ \\
		\bottomrule
	\end{tabular}
	\label{tbl:toy_prop}
\end{table}

\subsection{Network architecture and training parameters}
The detailed information on the network architecture and the training parameters for optimization of toy objective functions are summarized in Table~\ref{tbl:param_toy}.

\begin{table}[]
	\caption{Network architecture and training parameters for optimization of toy functions.}
	\centering
	\begin{tabular}{lcl}
		\toprule
		Description     & Symbol     & Value \\
		\midrule
		Number of samples drawn from the proposal distribution & $N$ & 20000\\
		Coefficient for the information capacity & $\gamma$ & 0.1\\
		Learning rate & & 0.001 \\
		Batch size & & 250 \\
		Number of training epoch && 350 \\
		Number of units in hidden layers in $q_{\vect{\psi}}(\vect{\xi}|z)$    &        & (64, 64)  \\
		Number of units in hidden layers in $p_{\vect{\theta}}(z|\vect{\xi})$    &        & (64, 64)  \\
		Activation function     &        & Relu, Relu  \\
		optimizer & & Adam\\
		\bottomrule
	\end{tabular}
	\label{tbl:param_toy}
\end{table}

\newpage

\section{Details of Motion Planning Tasks}

\subsection{Shaping function}
Given a set of samples drawn from a proposal distribution $\mathcal{D}=\{ \vect{x}_i \}^N_{i=1}$, we used the shaping function given by
\begin{align}
f(R(\vect{x})) = \exp\left( \frac{\alpha\big( R(\vect{x}) - R_{\textrm{max}} \big) }{ R_{\textrm{max}} - R_{\textrm{min}} } \right),
\end{align}
where $R_{\textrm{max}} = \max_{\vect{x} \in \mathcal{D}} R(\vect{x})$ and $R_{\textrm{min}} = \min_{\vect{x} \in \mathcal{D}} R(\vect{x})$, $a$ is a coefficient and $a = 20$ for learning the one-dimensional latent variable and $a=10$ for learning the two-dimensional latent variable.
For selecting the value of the hyperparameter $\alpha$, we investigated the performance with $\alpha =\{5, 10, 20, 50\}$.
When selecting the value of the hyperparameter $\alpha$, we observed a trade-off; if the value of $\alpha$ is low, the model $p_{\vect{\theta}}(\vect{\xi}|\vect{z})$ tends to model diverse behaviors, but the variance of the quality of the generated trajectories is high. When the value of $\alpha$ is high, the variance of the quality is low, but the diversity of the modeled behavior is relatively limited.
This effect of $\alpha$ is reasonable as the higher value of $\alpha$ leads to the sharper distribution of $p_{\textrm{target}}$ in (4) in the main text. 

\subsection{Objective function}
In the motion planning tasks, the objective function is $R(\vect{\xi}) = - \mathcal{C}(\vect{\xi})$, where $\mathcal{C}(\vect{\xi})$
is the objective function used in previous studies on trajectory optimization~\citep{Zucker13,Kalakrishnan11} given by  
\begin{align}
\mathcal{C}(\vect{\xi}) = c_{\textrm{obs} }(\vect{\xi}) + \alpha c_{\textrm{smoothenss} }(\vect{\xi}).
\label{eq:mp_obj}
\end{align}
The first term in \eqref{eq:mp_obj}, $c_{\textrm{obs} }(\vect{\xi})$, is the penalty for collision with obstacles.
Given a configuration $\vect{q}$, we denote by $\vect{x}_{u}(\vect{q}) \in \Real^3$ the position of the bodypoint $u$ in task space. $c_{\textrm{obs} }(\vect{\xi})$ is then given by
\begin{equation}
c_{\textrm{obs} }(\vect{\xi}) = \frac{1}{2} \sum_{t} \sum_{u \in \mathcal{B} } c \left( \vect{x}_{u}(\vect{q}_t) \right)  \left\| \frac{d}{dt}\vect{x}_{u}(\vect{q}_t) \right\|,
\end{equation}
and $\mathcal{B}$ is a set of body points that comprise the robot body. The local collision cost function $c(\vect{x}_{u})$ is defined as
\begin{equation}
c ( \vect{x}_{u} ) =
\left\{
\begin{array}{cll}
0 , & \textrm{if} & d(\vect{x}_{u}) >  \epsilon, \\
\frac{1}{2  \epsilon} (d (\vect{x}_{u}) -   \epsilon)^{2} , & \textrm{if} & 0 < d(\vect{x}_{u}) < \epsilon, \\
- d(\vect{x}_{u})+\frac{1}{2}\epsilon,& \textrm{if} & d(\vect{x}_{u})< 0,
\end{array}
\right. 
\end{equation}
where $\epsilon$ is the constant that defines the margin from the obstacle, and $d (\vect{x}_{u})$ is the shortest distance in task space between the bodypoint $u$ and obstacles.
The second term in \eqref{eq:mp_obj},  $c_{\textrm{smoothenss} }(\vect{\xi})$, is the penalty on the acceleration defined as $c_{\textrm{smoothenss} }(\vect{\xi}) = \sum^T_{t=1} \left\| \ddot{\vect{q}}_t \right\|^2.$

To make the computation efficient, the body of the robot manipulator and obstacles are approximated by a set of spheres. 

\subsection{Network architecture and training parameters}
The detailed information on the network architecture and the training parameters for motion planning tasks are summarized in Table~\ref{tbl:param_mp}.

\begin{table}[]
	\caption{Network architecture and training parameters for motion planning tasks.}
	\centering
	\begin{tabular}{lcl}
		\toprule
		Description     & Symbol     & Value \\
		\midrule
		Number of time steps in a trajectory & $T$ & 50\\
		Number of samples drawn from the proposal distribution & $N$ & 20000\\
		Coefficient for the information capacity & $\gamma$ & 10\\
		Learning rate & & 0.001 \\
		Batch size & & 250 \\
		Number of training epoch && 700 \\
		Number of units in hidden layers in $q_{\vect{\psi}}(\vect{\xi}|z)$    &        & (300, 200)  \\
		Number of units in hidden layers in $p_{\vect{\theta}}(z|\vect{\xi})$    &        & (200, 300)  \\
		Activation function     &        & Relu, Relu  \\
		optimizer & & Adam\\
		\bottomrule
	\end{tabular}
	\label{tbl:param_mp}
\end{table}

%

\subsection{Details on problem settings}
\begin{figure}
	\centering
	\begin{subfigure}[t]{0.3\columnwidth}
		\centering
		\includegraphics[width=\textwidth]{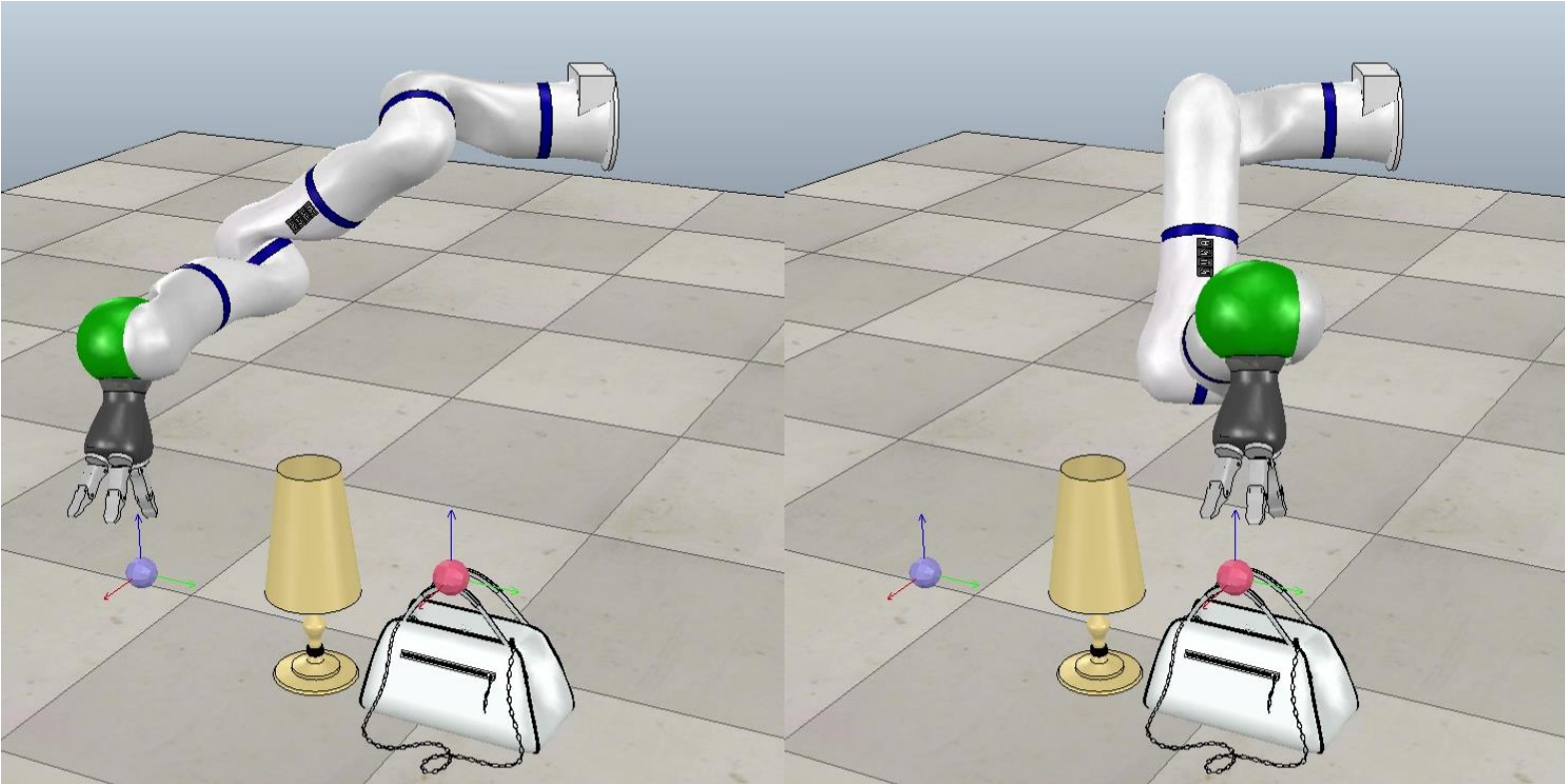}
		\caption{Problem setting of Task~1.  }
	\end{subfigure}
	\begin{subfigure}[t]{0.3\columnwidth}
		\centering
		\includegraphics[width=\textwidth]{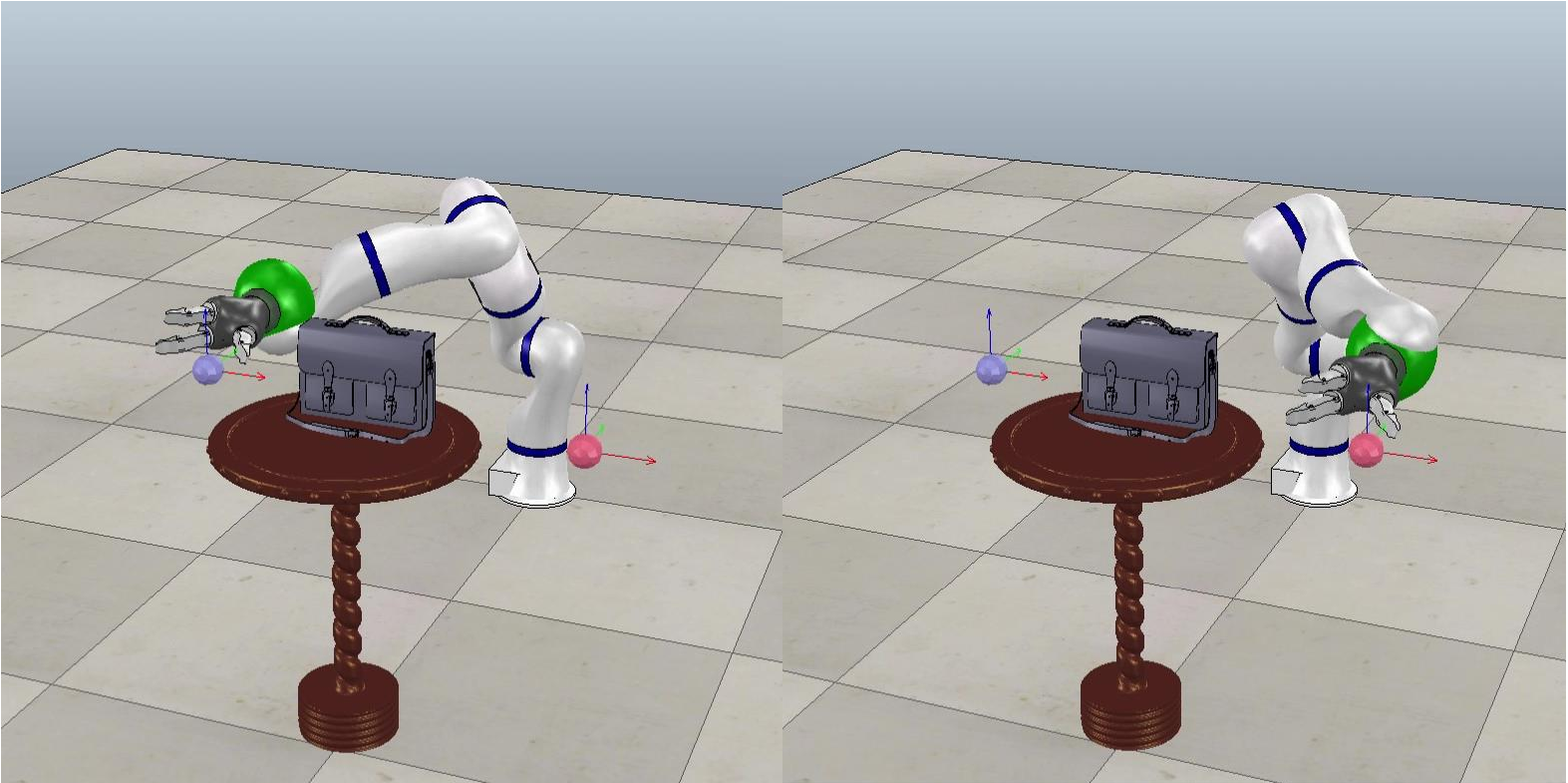}
		\caption{Problem setting of Task~2. }
	\end{subfigure}
	\begin{subfigure}[t]{0.3\columnwidth}
		\centering
		\includegraphics[width=\textwidth]{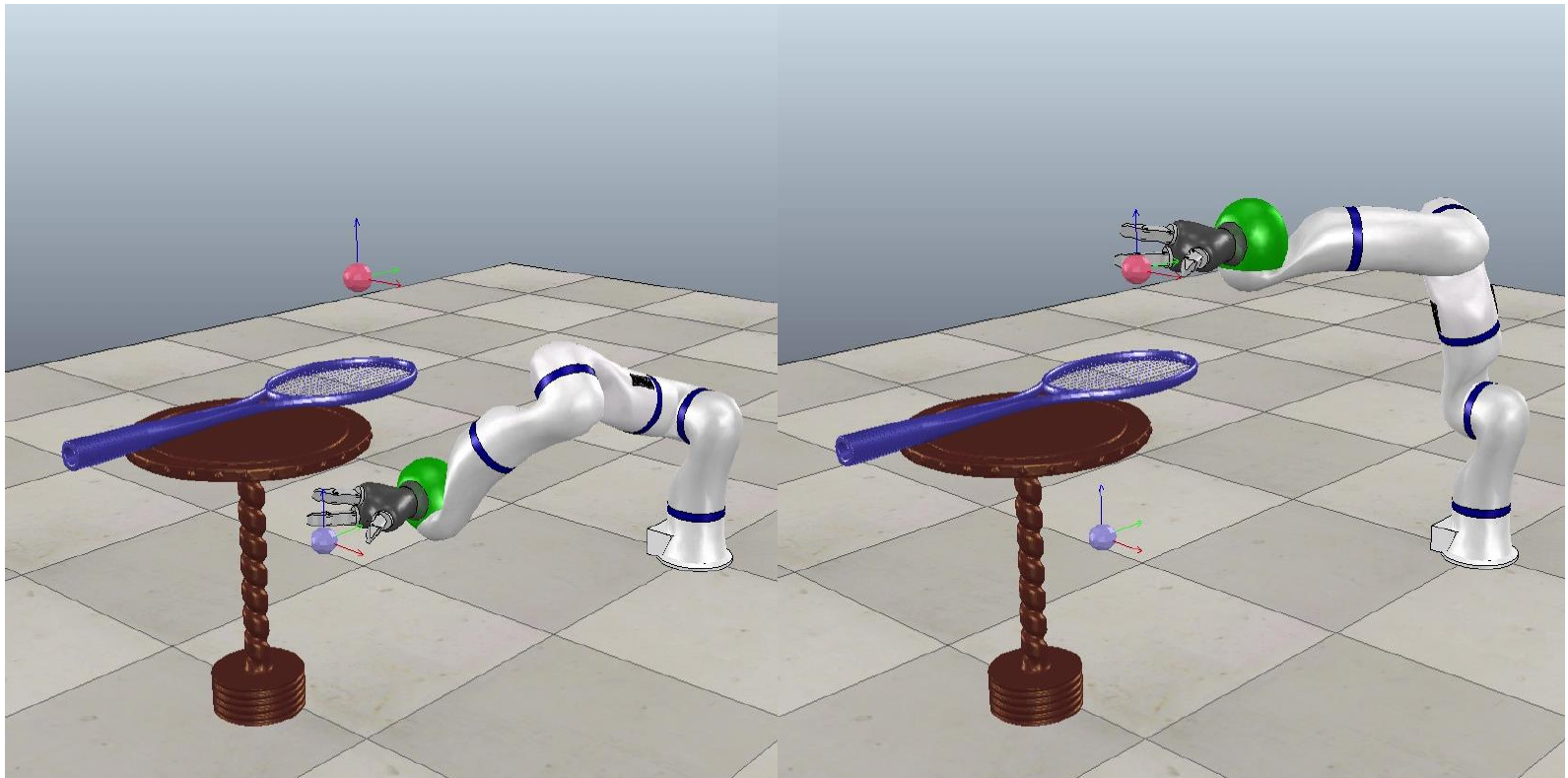}
		\caption{Problem setting of Task~3.  }
	\end{subfigure}
	
	\caption{Settings of motion planning tasks. Left: Start position. Right: goal position. 
		The blue sphere indicates the start position and the red sphere indicates the goal position.
	} 
	\label{fig:task_setting}
\end{figure}

The task setting of Task~1 is shown in Figure~\ref{fig:task_setting}(a). The blue sphere indicates the start position and the red sphere indicates the goal position.
The goal is to plan a trajectory to reach the goal position for grasping a bag, while avoiding a lamp.

The task setting of Task~2 is shown in Figure~\ref{fig:task_setting}(b). 
The goal is to plan a trajectory for reaching the goal position indicated by the red sphere, while avoiding a table and a bag.

The task setting of Task~3 is shown in Figure~\ref{fig:bat_loss_kl}(a). 
The goal is to plan a trajectory for reaching the goal position indicated by the red sphere, while avoiding a table and a tennis bat.

When we evaluate the trajectories generated with LSMO, we linearly interpolate $z$ in $[-1.28, 1.28]$ and generate seven samples from $p_{\vect{\theta}}(\vect{x}|\vect{z})$. 
As we run the experiment three times with different random seeds, we evaluated 21 samples in total and summarized the results in Table~2 in the main manuscript and Tables~\ref{tbl:dimension} and \ref{tbl:motion_time} in this supplementary.

\subsection{Effect of the dimensionality of the latent variable}
The appropriate dimensionality of the latent variable is dependent on tasks.
We show the result of learning one-dimensional and two-dimensional latent variables on Tasks~1, 2, and 3 in this section.
Although it is possible to iterate 1-4 in Algorithm~2 several times, we ran this procedure once to obtain the following results since we did not observe an improvement in the behavior after the second iteration.

\begin{figure}
	\centering
	\begin{subfigure}[t]{0.23\columnwidth}
		\centering
		\includegraphics[width=\textwidth]{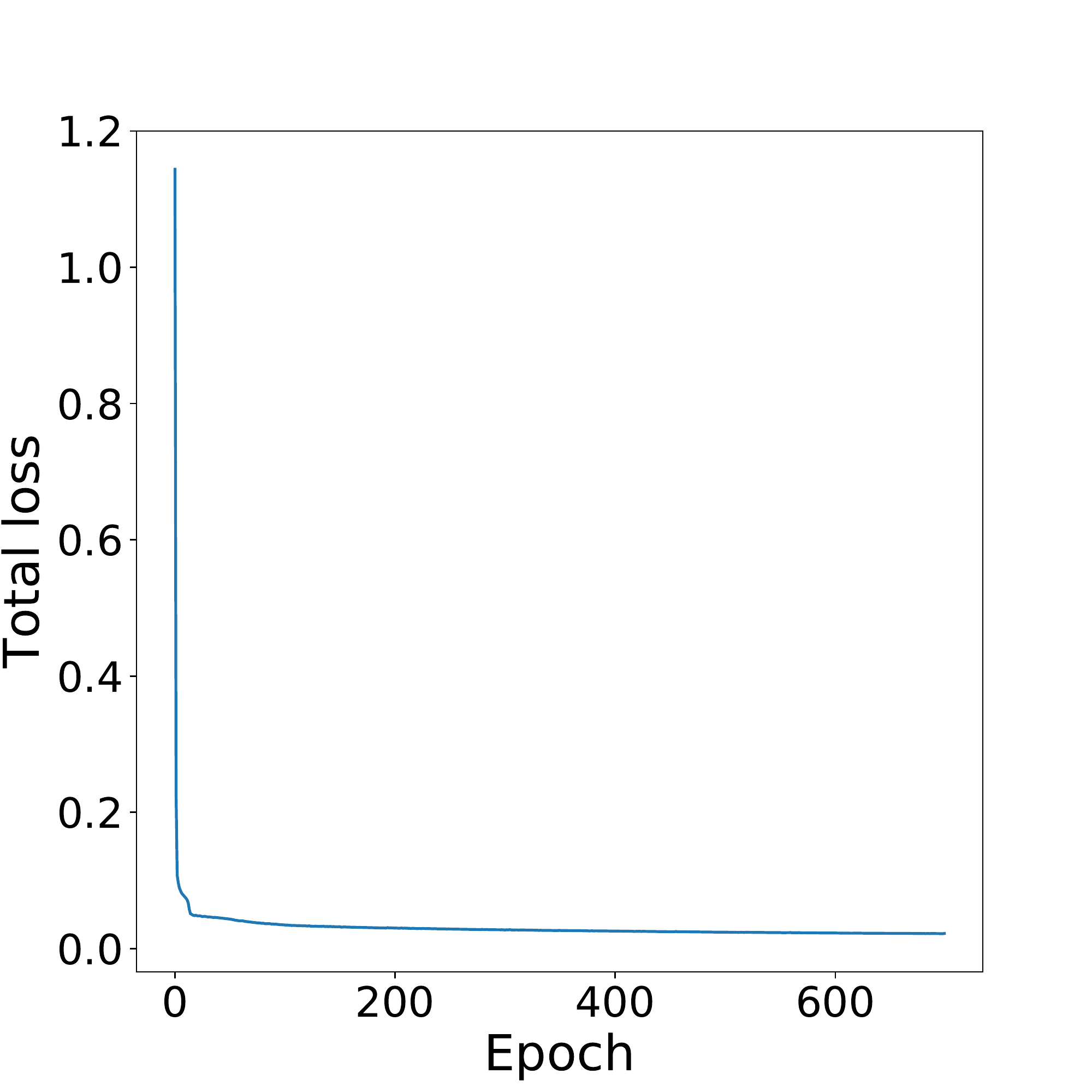}
		\caption{Values of loss function when $\vect{z}$ is two-dimensional. }
	\end{subfigure}
	\begin{subfigure}[t]{0.23\columnwidth}
		\centering
		\includegraphics[width=\textwidth]{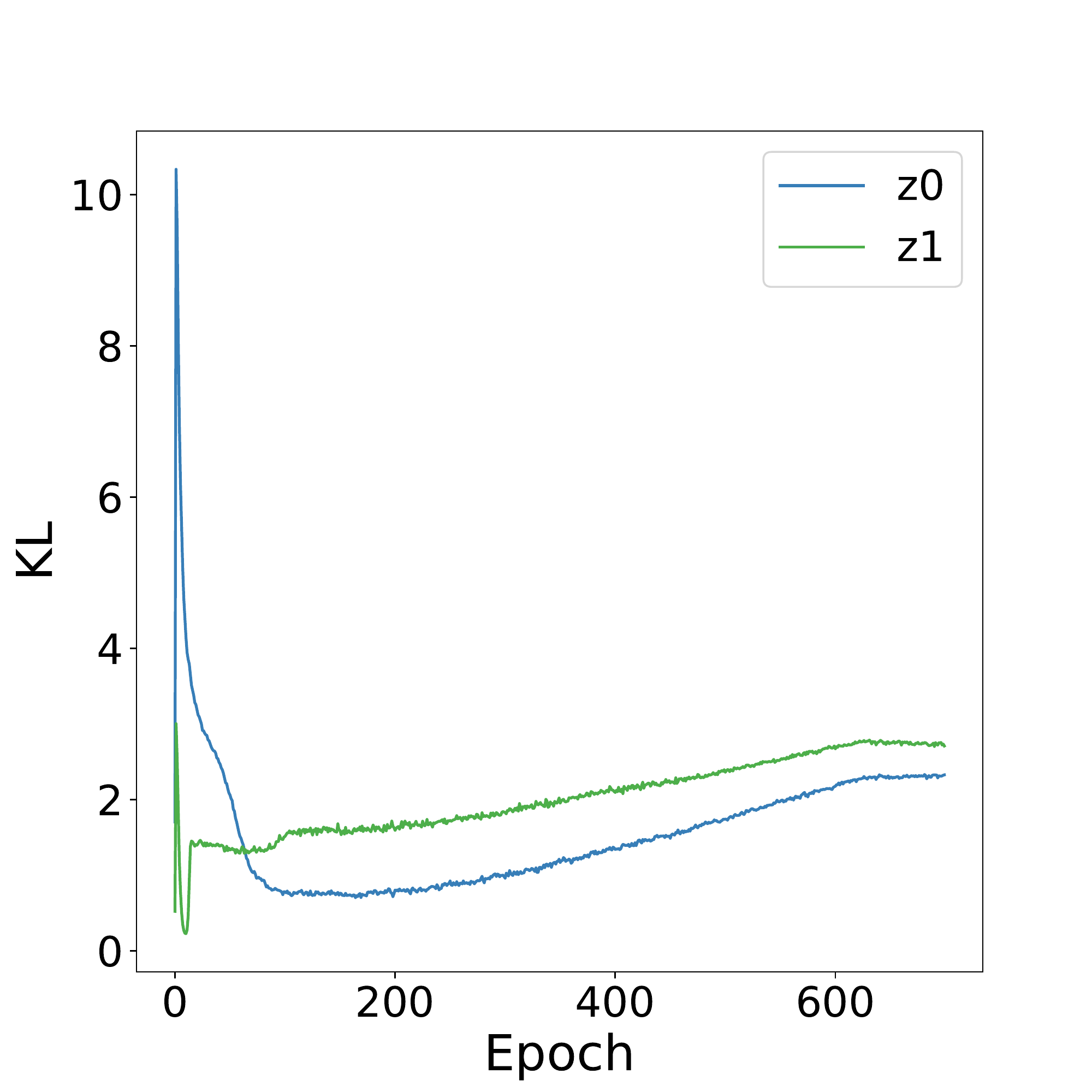}
		\caption{$\KL\big( q(z | \vect{\xi}) || p(z) \big)$ when $\vect{z}$ is two-dimensional. } 
	\end{subfigure}
	\begin{subfigure}[t]{0.23\columnwidth}
		\centering
		\includegraphics[width=\textwidth]{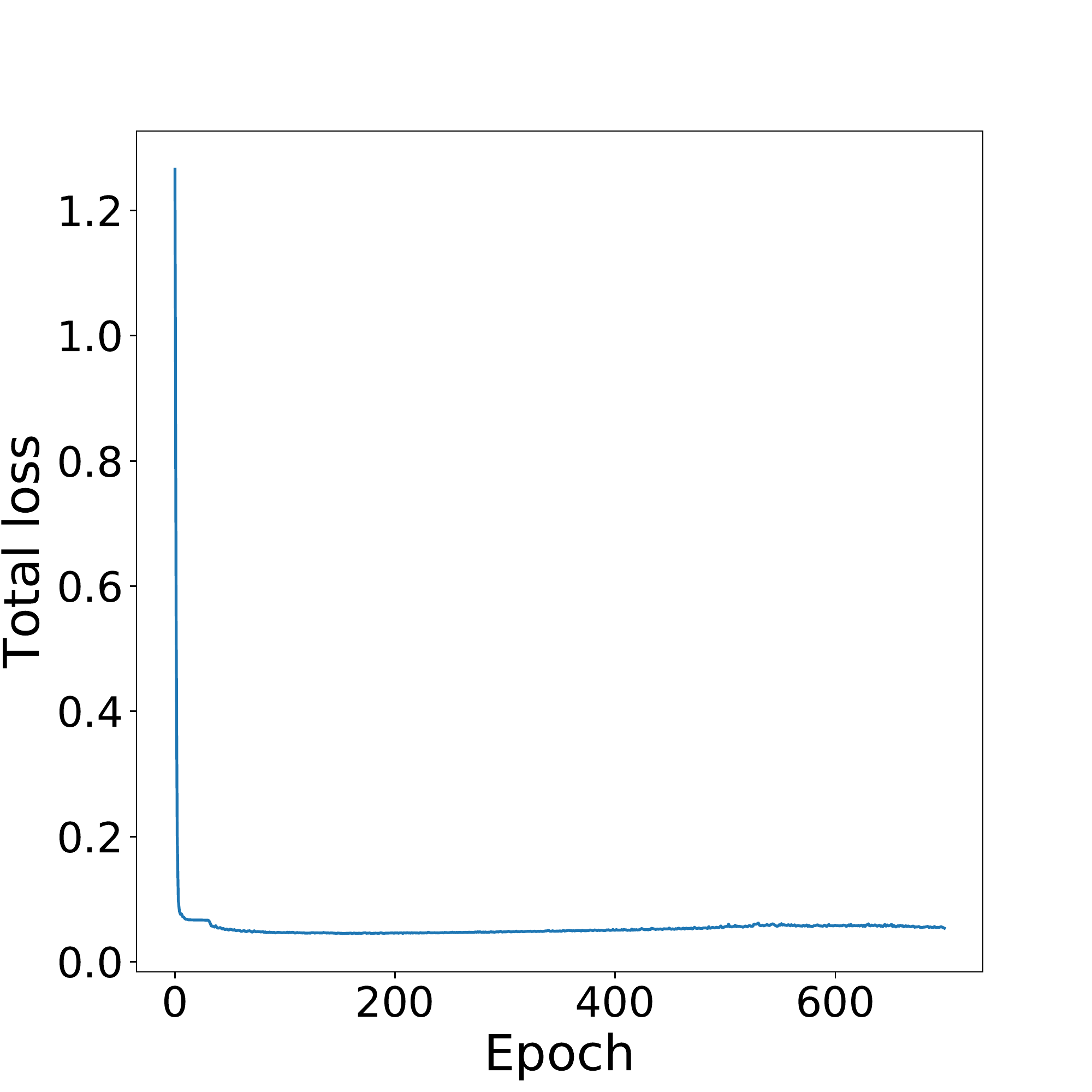}
		\caption{Values of the loss function when $\vect{z}$ is one-dimensional. }
	\end{subfigure}
	\begin{subfigure}[t]{0.23\columnwidth}
		\centering
		\includegraphics[width=\textwidth]{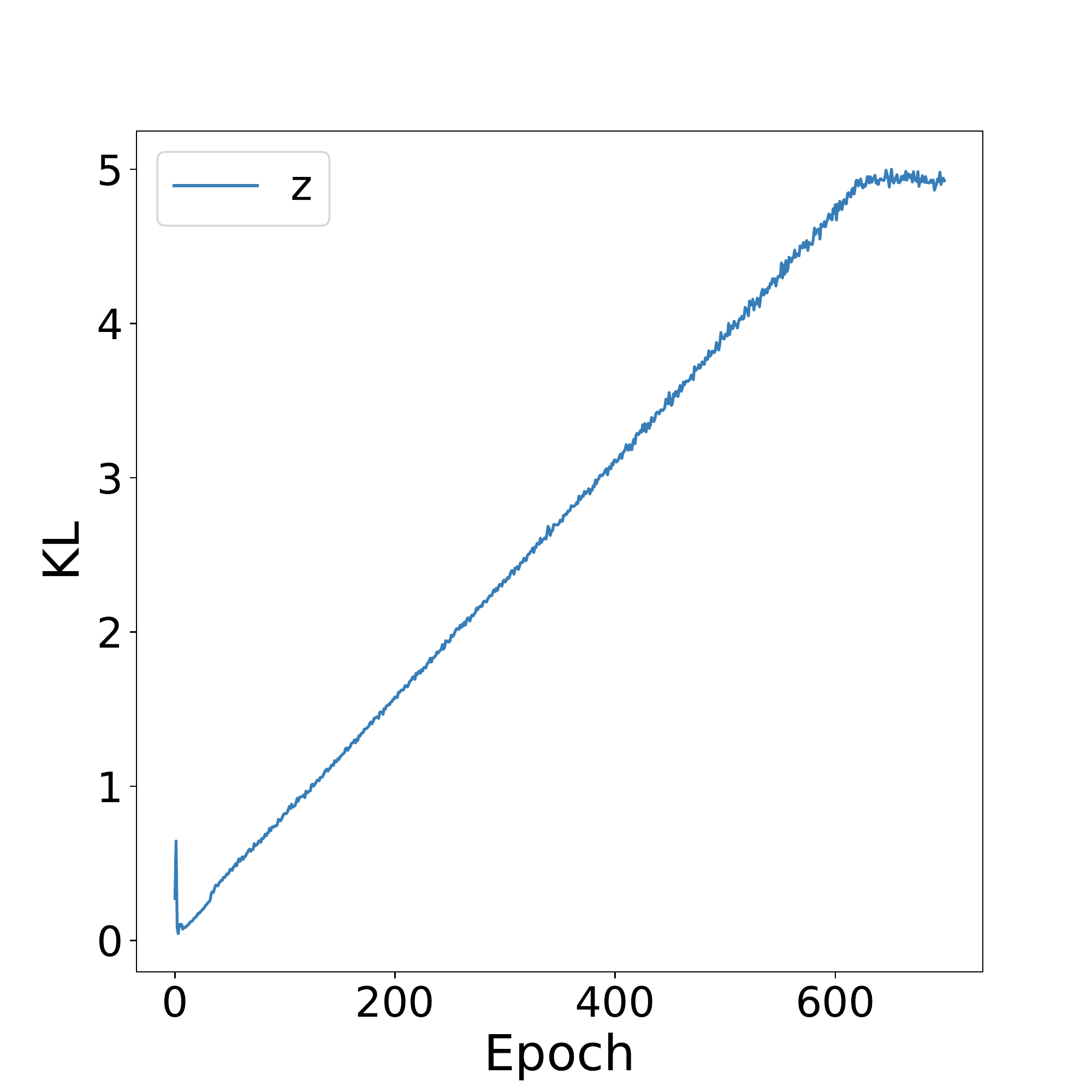}
		\caption{$\KL\big( q(z | \vect{\xi}) || p(z) \big)$  when $\vect{z}$ is one-dimensional. } 
	\end{subfigure}
	\caption{Loss function and the KL divergence during training for Task~1. (a) and (b): Training when  $\vect{z}$ is two-dimensional. (c) and (d): Training when  $\vect{z}$ is one-dimensional. 
	} 
	\label{fig:bag_loss_kl}
\end{figure}

\begin{figure}
	\begin{subfigure}[t]{\columnwidth}
		\includegraphics[width=\textwidth]{LSMO_bag_2}
		\caption{The latent variable $\vect{z}$ is two-dimensional as $\vect{z} = [z_0, z_1]$ in this result. The top row shows the variation with different $z_0$ and the second row shows the variation with different $z_1$. Trajectories indicated by the blue square show the trajectory generated with $z_0=z_1 =0$.} 
	\end{subfigure}
	\begin{subfigure}[t]{\columnwidth}
		\includegraphics[width=\textwidth]{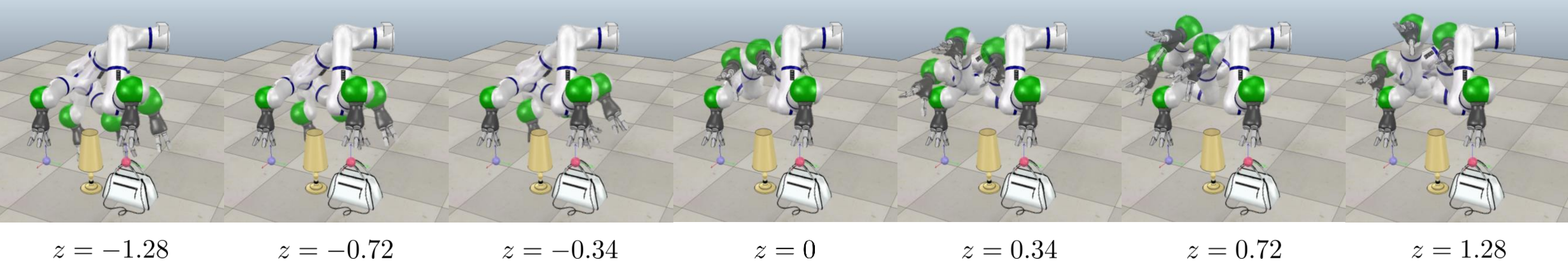}
		\caption{The latent variable $\vect{z}$ is one-dimensional in this result. } 
	\end{subfigure}
	\caption{Visualization of solution obtained by LSMO on Task~1. Comparison of the results of the two-dimensional and one-dimensional latent variable.}
	\label{fig:bag}
\end{figure}

The values of the loss function and the KL divergence $\KL\big( q(\vect{z} | \vect{\xi}) || p(z) \big)$ during training for Task~1 are shown in Figure~\ref{fig:bag_loss_kl}. 
The plots in Figure~\ref{fig:bag_loss_kl}(b) indicate that the information encoded in the two channels was stably increased during the training when learning the two-dimensional latent variable. 
Figure~\ref{fig:bag} indicates that different variations of trajectories are encoded in the two channels.
As shown in the first row of Figure~\ref{fig:bag}(a), varying the value of $z_0$ leads to the change of the orientation of the end-effector while avoiding the obstacles. 
Meanwhile, varying the value of $z_1$ leads to the variation of the positions for avoiding the obstacle;
The end-effector goes over the obstacle if $z_1=-1.28$, and the end-effector goes behind the obstacle if $z_1=1.28$. 
When learning the latent variable is one-dimensional, the variation of the trajectories is limited compared to the result with the two-dimensional latent variable.

\begin{figure}[tb]
	\centering
	\begin{subfigure}[t]{0.23\columnwidth}
		\centering
		\includegraphics[width=\textwidth]{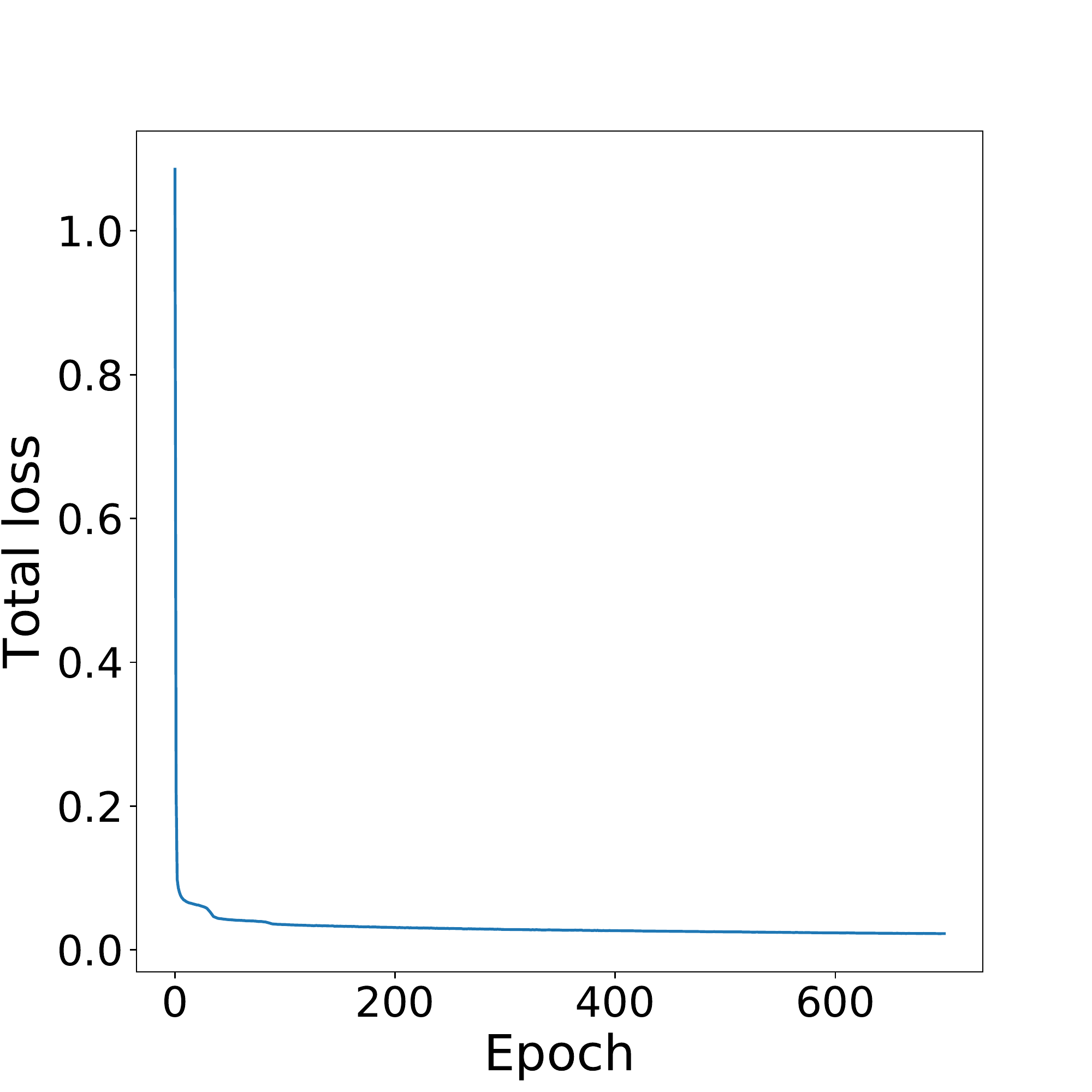}
		\caption{Values of loss function when $\vect{z}$ is two-dimensional. }
	\end{subfigure}
	\begin{subfigure}[t]{0.23\columnwidth}
		\centering
		\includegraphics[width=\textwidth]{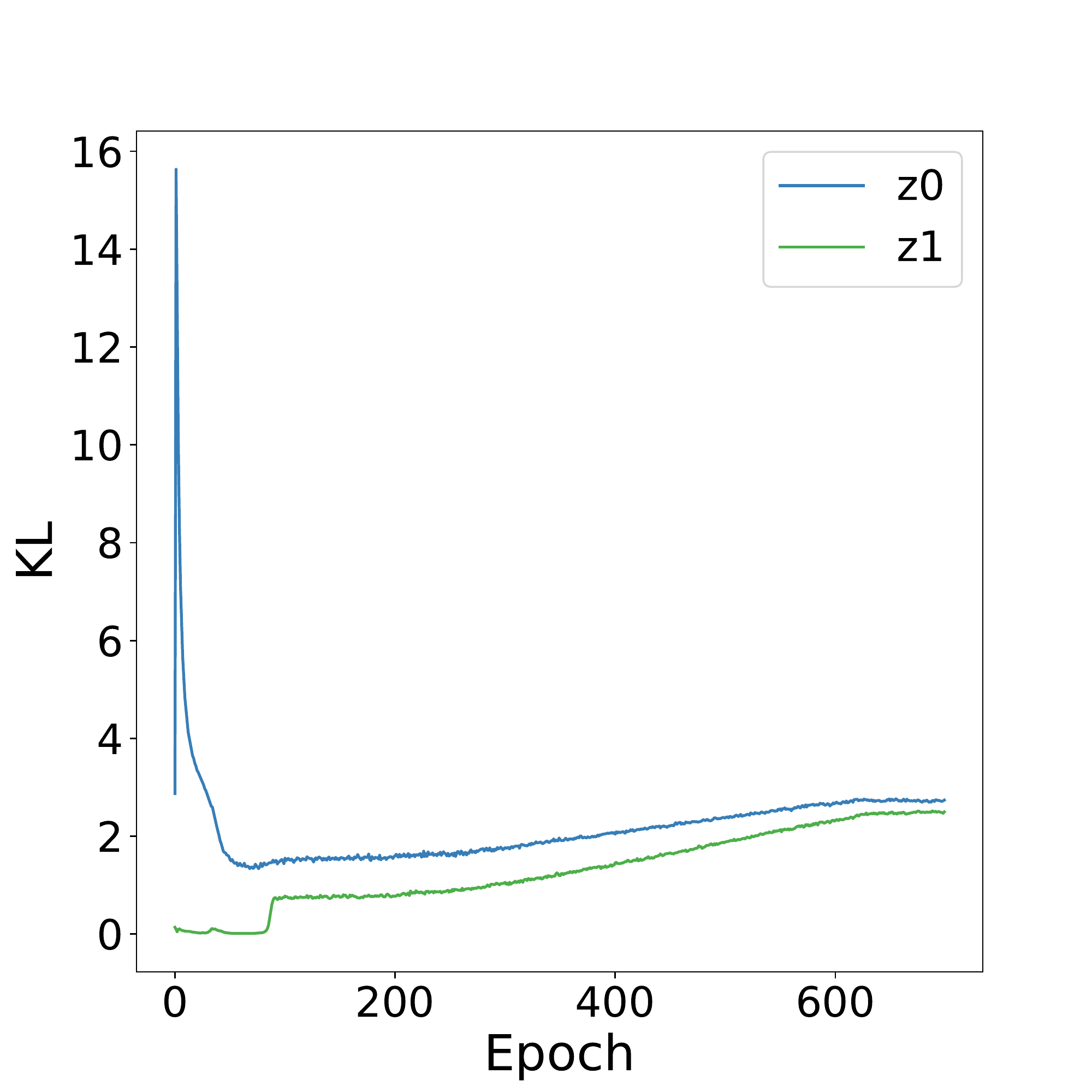}
		\caption{$\KL\big( q(z | \vect{\xi}) || p(z) \big)$ when $\vect{z}$ is two-dimensional. } 
	\end{subfigure}
	\begin{subfigure}[t]{0.23\columnwidth}
		\centering
		\includegraphics[width=\textwidth]{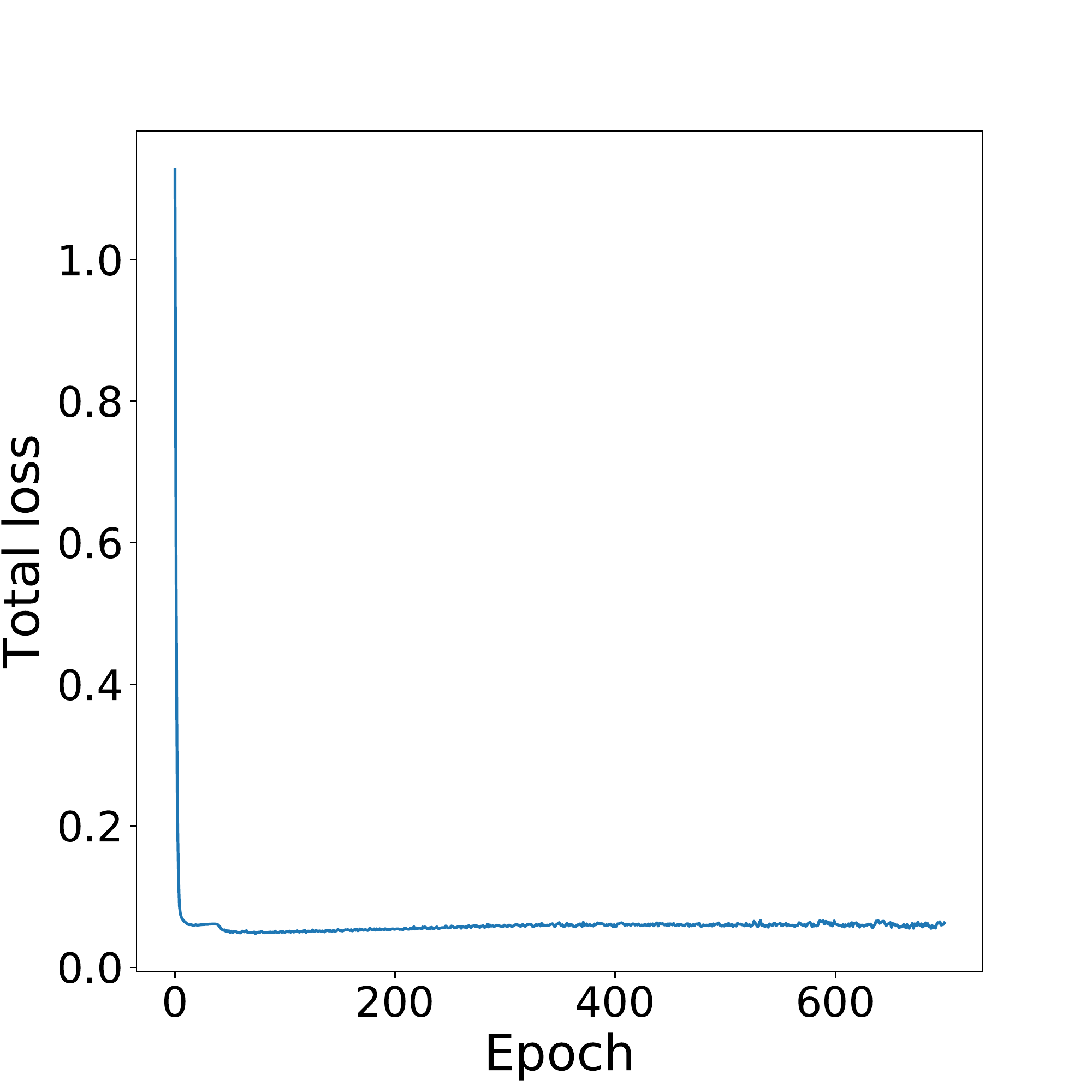}
		\caption{The values of the loss function. }
	\end{subfigure}
	\begin{subfigure}[t]{0.23\columnwidth}
		\centering
		\includegraphics[width=\textwidth]{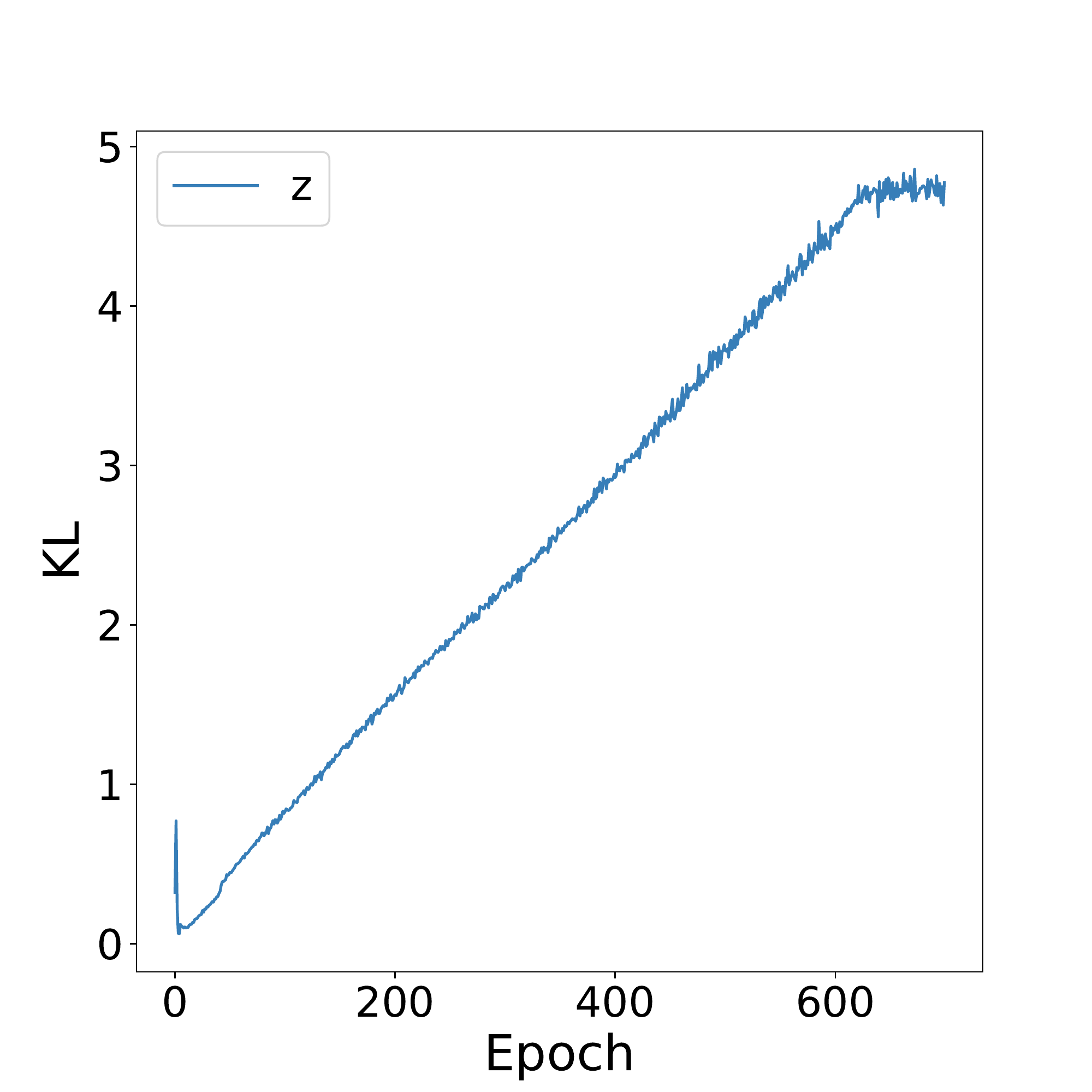}
		\caption{$\KL\big( q(z | \vect{\xi}) || p(z) \big)$. } 
	\end{subfigure}
	\caption{Loss function and KL divergence during training for Task~2. (a) and (b): Training when  $\vect{z}$ is two-dimensional. (c) and (d): Training when  $\vect{z}$ is one-dimensional. 
	} 
	\label{fig:table_loss_kl}
\end{figure}

\begin{figure}[tb]
	\begin{subfigure}[t]{\columnwidth}
		\includegraphics[width=\textwidth]{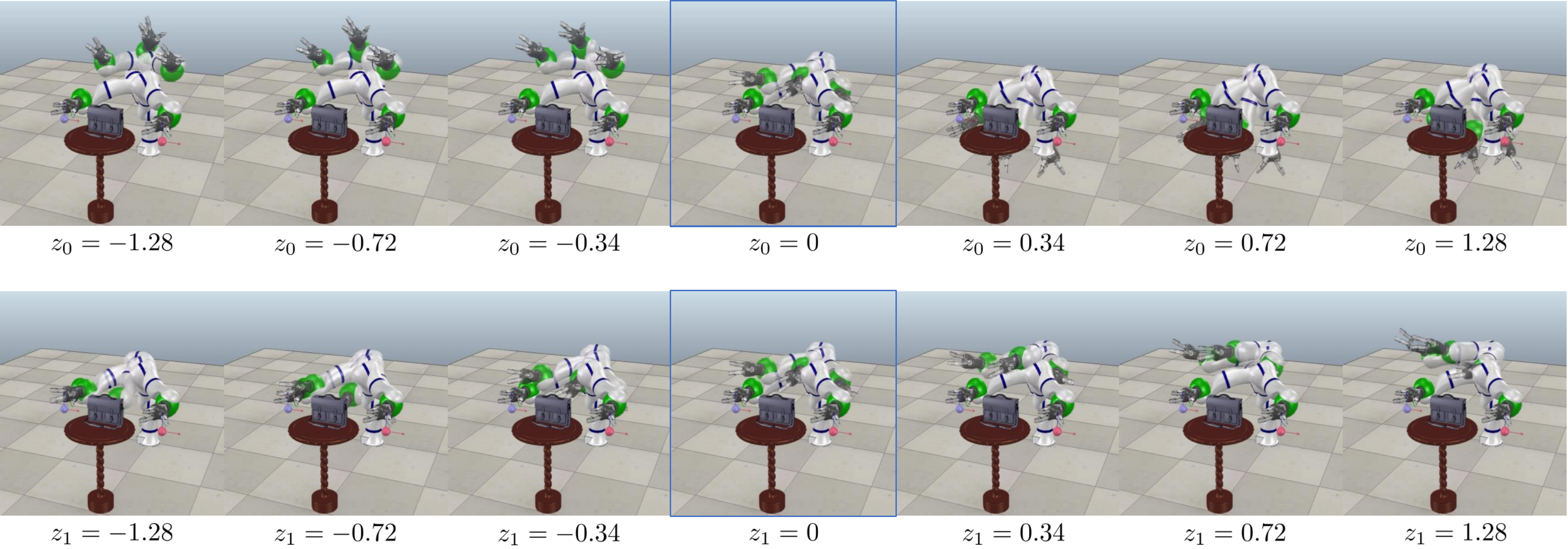}
		\caption{The latent variable $\vect{z}$ is two-dimensional as $\vect{z} = [z_0, z_1]$ in this result. The top row shows the variation with different $z_0$ and the second row shows the variation with different $z_1$. Trajectories indicated by the blue square show the trajectory generated with $z_0=z_1 =0$.} 
	\end{subfigure}
	\begin{subfigure}[t]{\columnwidth}
		\includegraphics[width=\textwidth]{LSMO_table}
		\caption{The latent variable $\vect{z}$ is one-dimensional in this result. } 
	\end{subfigure}
	\caption{Visualization of solution obtained by LSMO on Task~2. Comparison of the results with the two-dimensional and one-dimensional latent variables.}
	\label{fig:table}
\end{figure}

The values of the loss function and KL divergence $\KL\big( q(z | \vect{\xi}) || p(z) \big)$ during training for Task~2 is plotted in Figure~\ref{fig:table_loss_kl}.
Figure~\ref{fig:table_loss_kl}~(b) indicates that channel $z_1$ does not have much information compared with channel $z_0$.
If $z_0=-1.28$, the end-effector goes over the obstacles, and the end-effector goes under the obstacle if $z_0=1.28$.
Although the channel $z_1$ also encodes the height of the end-effector, the visualization of the variation of outputs shown in Figure~\ref{fig:table} indicates that variation of the channel $z_1$ does not induce much variation in the generated trajectories. 
Therefore, the information encoded in two channels is entangled on Task~2, and the results indicate that learning the one-dimensional latent variable is sufficient for Task~2.

\begin{figure}[tb]
	\centering
	\begin{subfigure}[t]{0.23\columnwidth}
		\centering
		\includegraphics[width=\textwidth]{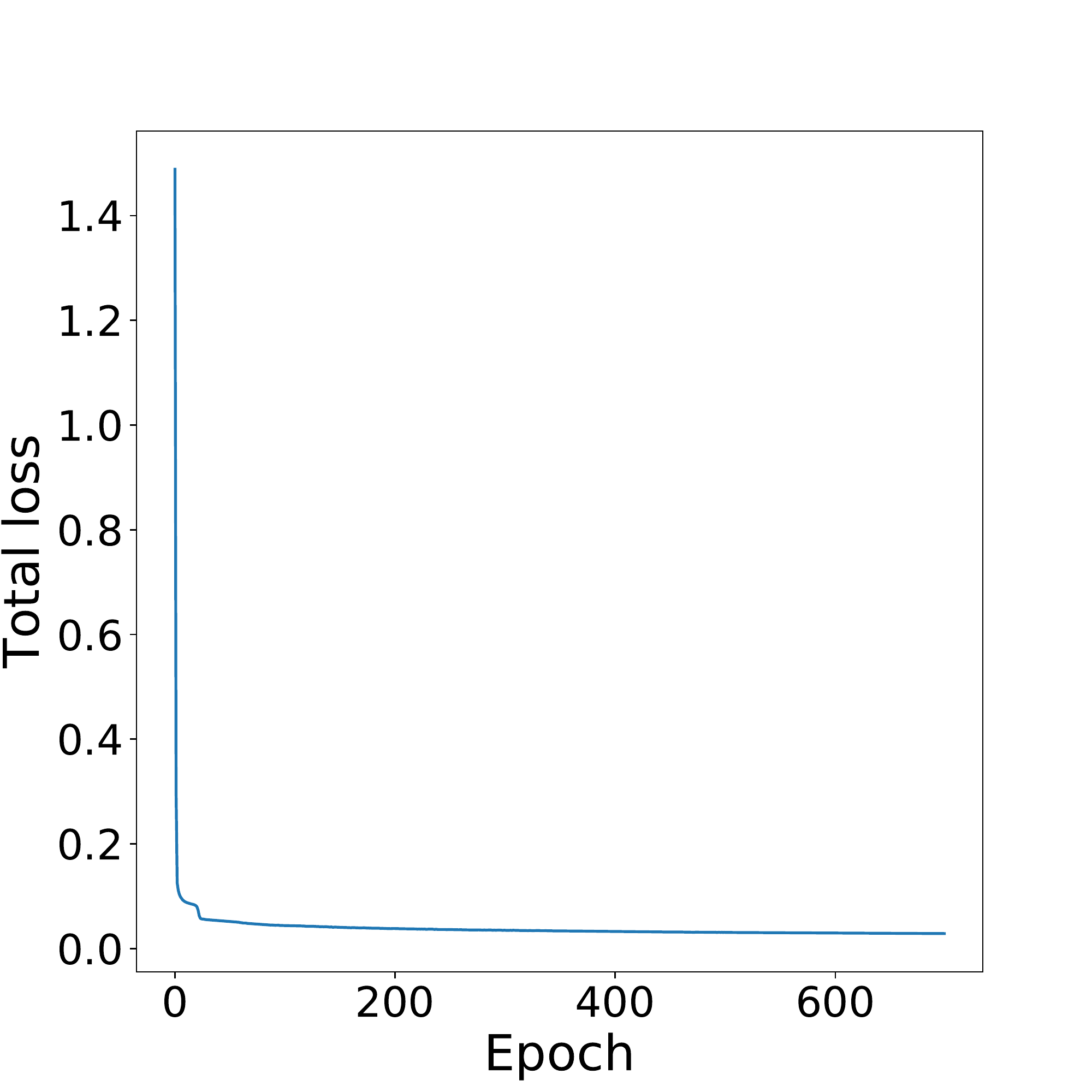}
		\caption{The values of the loss function when $\vect{z}$ is two-dimensional. }
	\end{subfigure}
	\begin{subfigure}[t]{0.23\columnwidth}
		\centering
		\includegraphics[width=\textwidth]{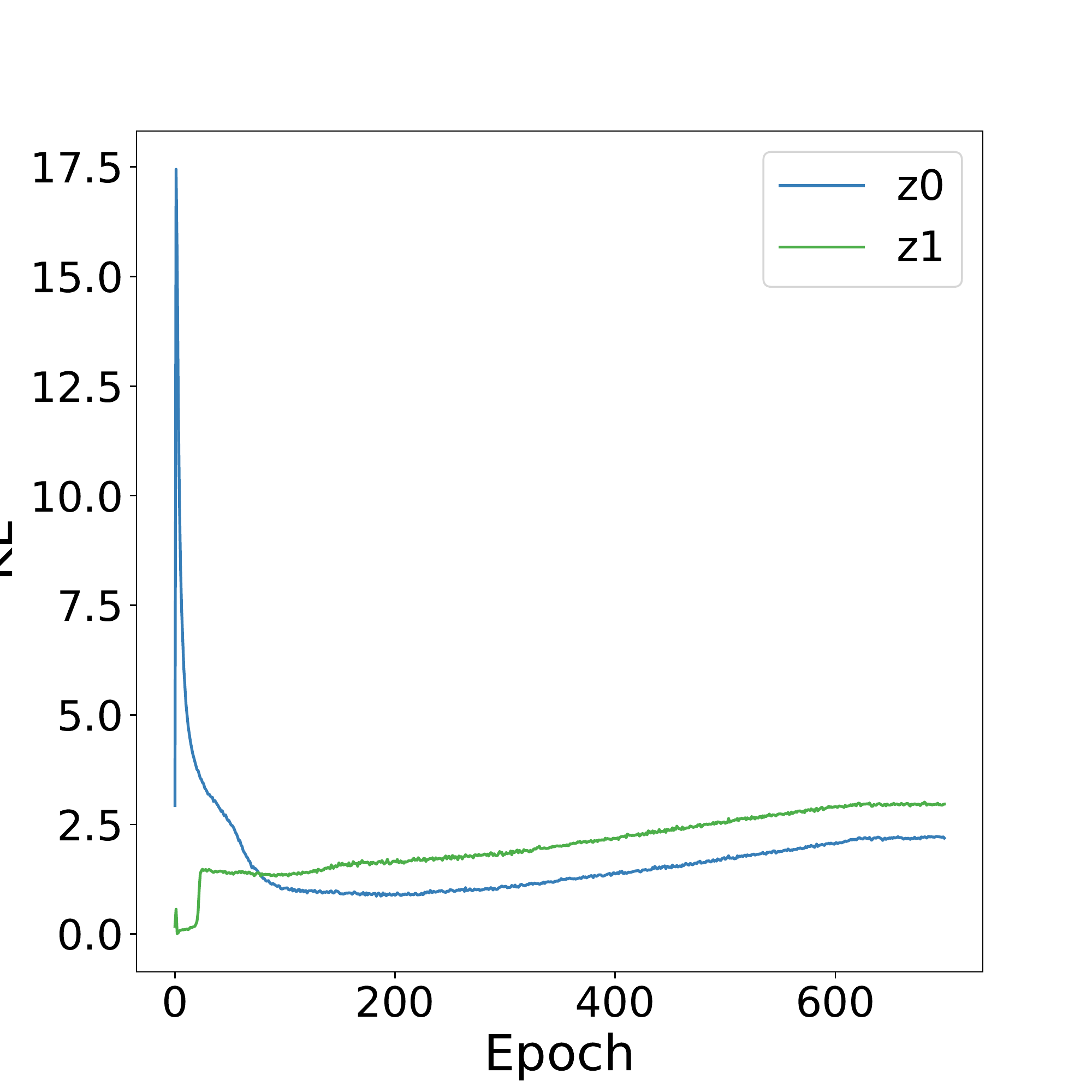}
		\caption{$\KL\big( q(z | \vect{\xi}) || p(z) \big)$ when $\vect{z}$ is two-dimensional. } 
	\end{subfigure}
	\begin{subfigure}[t]{0.23\columnwidth}
		\centering
		\includegraphics[width=\textwidth]{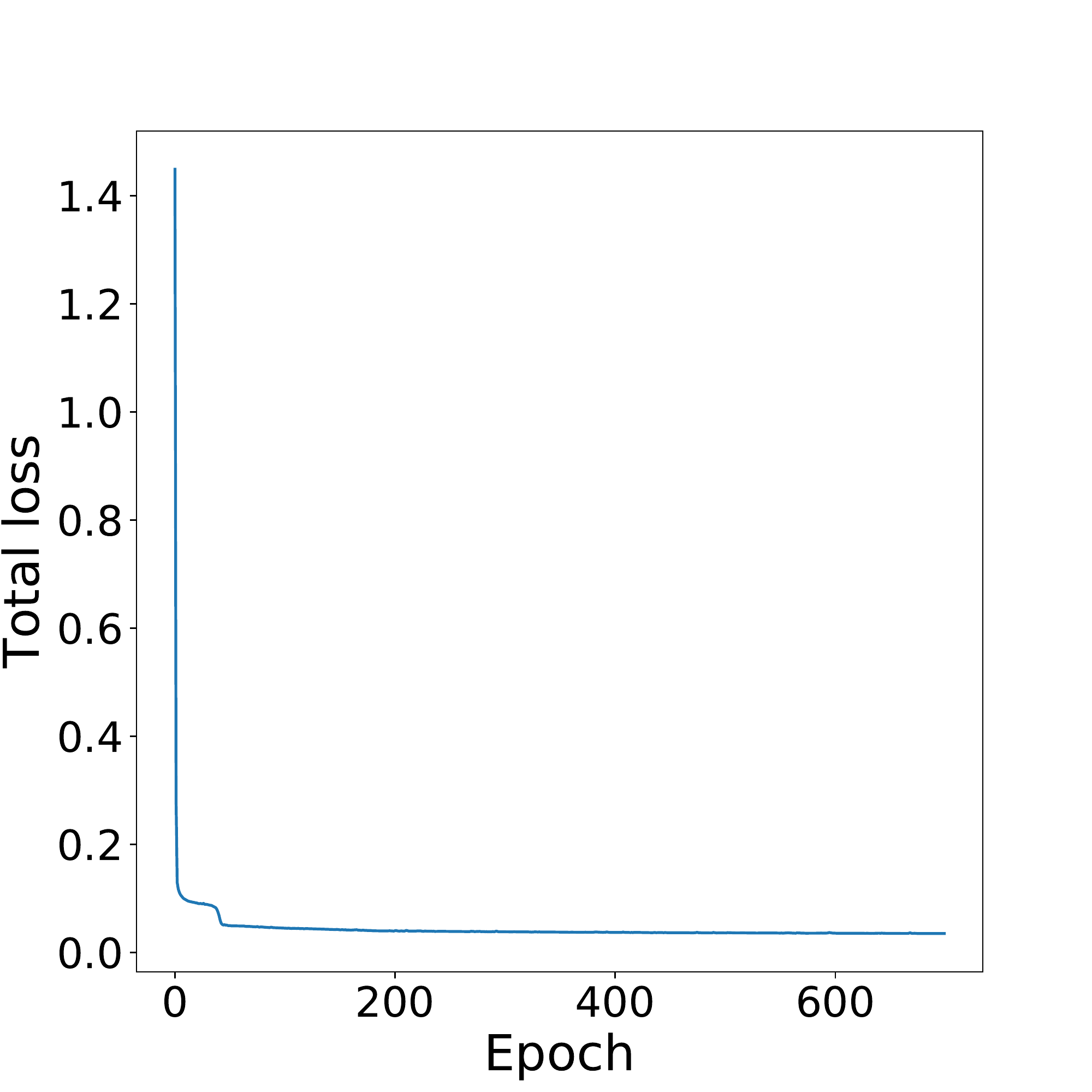}
		\caption{The values of the loss function. }
	\end{subfigure}
	\begin{subfigure}[t]{0.23\columnwidth}
		\centering
		\includegraphics[width=\textwidth]{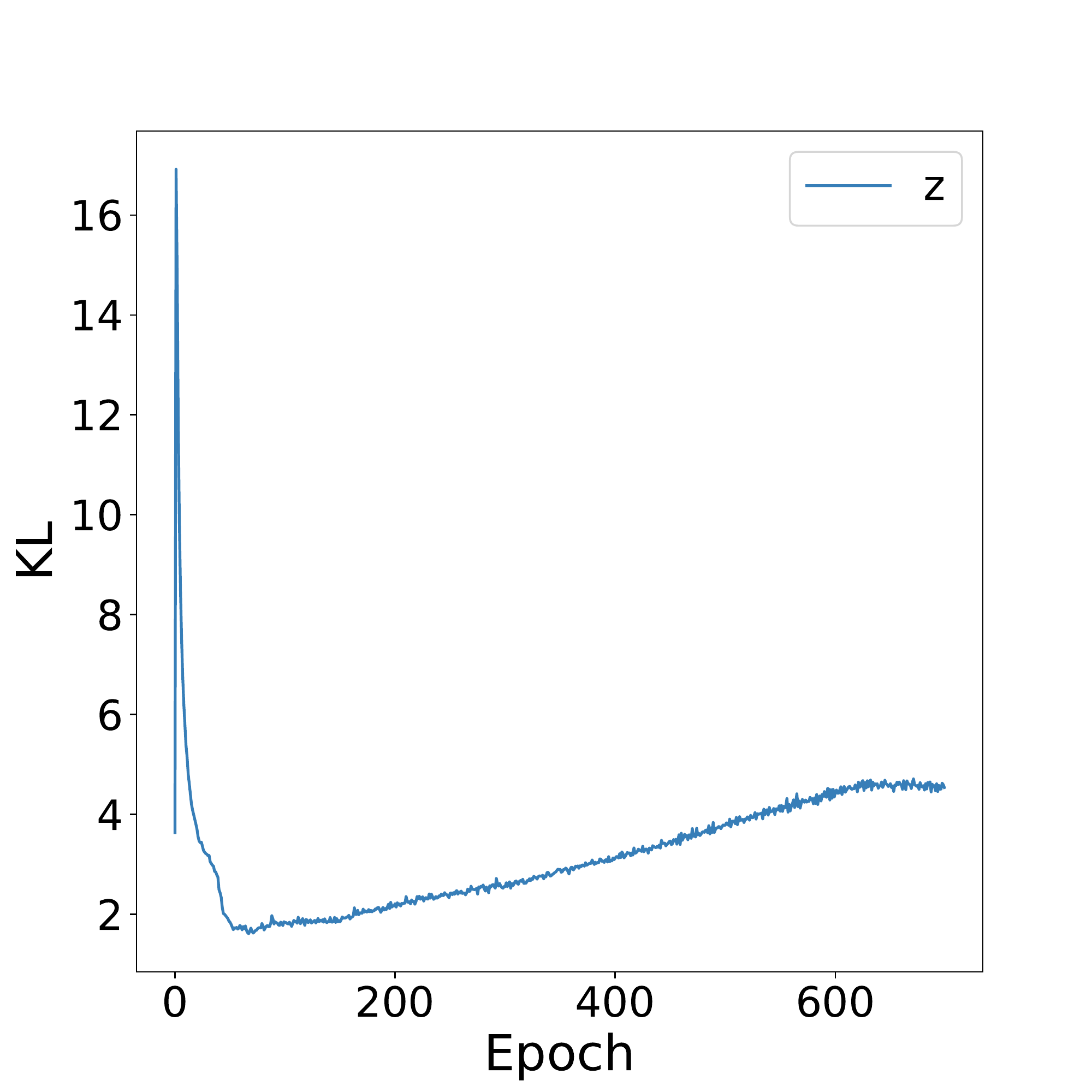}
		\caption{$\KL\big( q(z | \vect{\xi}) || p(z) \big)$. } 
	\end{subfigure}
	\caption{The loss function and the KL divergence during the training for Task~3. (a) and (b) show the training when  $\vect{z}$ is two-dimensional. (c) and (d) show the training when  $\vect{z}$ is one-dimensional.
	} 
	\label{fig:bat_loss_kl}
\end{figure}

\begin{figure}[tb]
	\begin{subfigure}[t]{\columnwidth}
		\includegraphics[width=\textwidth]{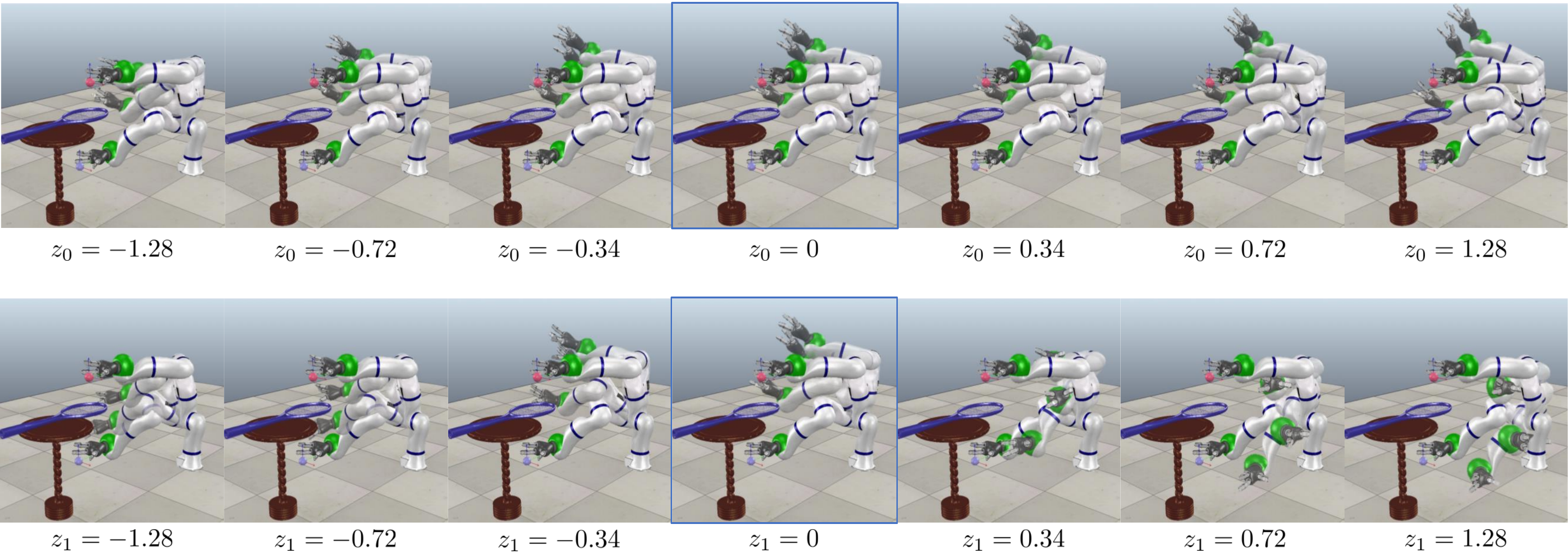}
		\caption{The latent variable $\vect{z}$ is two-dimensional as $\vect{z} = [z_0, z_1]$ in this result. The top row shows the variation with different $z_0$ and the second row shows the variation with different $z_1$. Trajectories indicated by the blue square show the trajectory generated with $z_0=z_1 =0$.} 
	\end{subfigure}
	\begin{subfigure}[t]{\columnwidth}
		\includegraphics[width=\textwidth]{LSMO_bat}
		\caption{The latent variable $\vect{z}$ is one-dimensional in this result. } 
	\end{subfigure}
	\caption{Visualization of solution obtained by LSMO on Task~3. Comparison of the results of the two-dimensional and one-dimensional latent variables.}
	\label{fig:bat}
\end{figure}

The values of the loss function and KL divergence $\KL\big( q(z | \vect{\xi}) || p(z) \big)$ during training for Task~3 are plotted in Figure~\ref{fig:bat_loss_kl}.
Figure~\ref{fig:bat_loss_kl}(b) indicates that the channels $z_1$ encodes more significant information than the channel $z_0$.
From Figure~\ref{fig:bat}(a), it is evident that the variation of $z_0$ and $z_1$ leads to different variations of the generated trajectories.
When varying the value of $z_0$, the height of the end-effector during the motion changes.
Meanwhile, when varying the value of $z_1$, positions for avoiding the obstacles change as shown in the second row of Figure~\ref{fig:bat}(a); When $z_1=-1.28$, the end-effector passes the left-hand side of the obstacle, while the end-effector passes the right-hand side when $z_1=1.28$.
Therefore, the information is disentangled in these two channels.

The scores of the trajectories obtained by the model with the 1d and 2d latent variables are summarized in Table~\ref{tbl:dimension}.
Scores are computed for trajectories generated without fine-tuning. 
When learning the two-dimensional latent variable, the variance of the score is much larger than that of the results obtained with the one-dimensional latent variable.
This result indicates that the one-dimensional latent variable is sufficient to represent the distribution of optimal points on these motion-planning tasks.

\begin{table}[tb]
	\vspace{-0.5cm}
	\caption{Comparison of scores between the one-dimensional and two-dimensional latent variables (higher is better).
	}
	\centering
	\begin{tabular}{llll}
		\toprule
		& Task~1     & Task~2  & Task~3	 \\
		\midrule
		LSMO (1d) & $-2.95 \pm 1.33 $  & $ -2.24 \pm 1.15$ & $-3.24 \pm 0.92 $  \\
		LSMO (2d) &  $-4.37 \pm 2.44 $  & $ -4.10 \pm 3.00$ & $-4.87 \pm 2.48 $\\
		\bottomrule
	\end{tabular}
	\label{tbl:dimension}
\end{table}

\section{Relation between the motion planning problems and RL problems}
The formulation of RL is based on the Markov decision process~(MDP)~\cite{Sutton98}, and the focus of RL is to learn a policy that maximizes the expected return in a stochastic environment. In addition, the learning agent is often an underactuated robot.
In contrast, the target of the motion-planning problem addressed in this study is to obtain a trajectory that minimizes the cost function.
In the context of motion planning, a trajectory is a sequence of robot configurations.
The formulation of the motion-planning problem is based on the assumption that the robotic system is fully actuated and lower-level controllers are available. 
This assumption is valid in many industrial robotic systems, which are rigid and powerful enough to move quickly in a factory.
For these reasons, trajectory optimization has been investigated for decades~\cite{Khatib86,Zucker13,Schulman14,Mukadam18}.
Because of the difference in the formulation, the methods of solving RL problems and the motion-planning problems are fairly different.
The challenge of trajectory optimization for the motion-planning problem involves dealing with a non-convex objective function that involves hundreds of parameters. 
Although previous studies addressed how to formulate the problem in a tractable manner, existing methods are designed to find a local-optima.
To the best of our knowledge, our method is the first approach that captures a set of homotopic solutions for a motion-planning problem.

\bibliographystyle{plainnat}
\bibliography{neurips20}

\end{document}